\documentclass[journal]{IEEEtran}
\usepackage{graphicx}
\usepackage{epsfig}
\usepackage{epstopdf}
\graphicspath {{/Figs}}
\usepackage{subcaption}
\usepackage{caption}
\usepackage{xcolor,colortbl}
\usepackage{ulem}

\usepackage[pagebackref=true,breaklinks=true,letterpaper=true,colorlinks,bookmarks=false]{hyperref}

\ifCLASSINFOpdf
  
\else
  
\fi

\begin{document}

\title{Adaptive Subsampling for ROI-based Visual Tracking: Algorithms and FPGA Implementation}

\author{Odrika Iqbal*,
        Victor Isaac Torres Muro*,
        Sameeksha Katoch,
        Andreas Spanias,~\IEEEmembership{Fellow,~IEEE,}
        and Suren Jayasuriya,~\IEEEmembership{Member,~IEEE}
\thanks{*Authors had equal contribution.}
}


\maketitle

\begin{abstract}
There is tremendous scope for improving the energy efficiency of embedded vision systems by incorporating programmable region-of-interest (ROI) readout in the image sensor design. In this work, we study how ROI programmability can be leveraged for tracking applications by anticipating where the ROI will be located in future frames and switching pixels off outside of this region. We refer to this process of ROI prediction and corresponding sensor configuration as adaptive subsampling. Our adaptive subsampling algorithms comprise an object detector and an ROI predictor (Kalman filter) which operate in conjunction to optimize the energy efficiency of the vision pipeline with the end task being object tracking. To further facilitate the implementation of our adaptive algorithms in real life, we select a candidate algorithm and map it onto an FPGA. Leveraging Xilinx Vitis AI tools, we designed and accelerated a YOLO object detector-based adaptive subsampling algorithm. In order to further improve the algorithm post-deployment, we evaluated several competing baselines on the OTB100 and LaSOT datasets. We found that coupling the ECO tracker with the Kalman filter has a competitive AUC score of 0.4568 and 0.3471 on the OTB100 and LaSOT datasets respectively. Further, the power efficiency of this algorithm is on par with, and in a couple of instances superior to, the other baselines. The ECO-based algorithm incurs a power consumption of approximately 4 W averaged across both datasets while the YOLO-based approach requires power consumption of approximately 6 W (as per our power consumption model). In terms of accuracy-latency tradeoff, the ECO-based algorithm provides near-real-time performance (19.23 FPS) while managing to attain competitive tracking precision. 
\end{abstract}


\IEEEpeerreviewmaketitle

\section{Introduction}
\vspace{-0.2cm}

There is a wide array of computer vision applications that feature object detection and tracking at their core~\cite{mocanu2018deep,chen2019real,cucchiara2002sakbot,dedeouglu2004moving,arnold2019survey,park2008multiple}. 
Surveillance, autonomous driving, drone navigation are among a myriad of applications that demand low-latency and high-precision tracking. Recent developments in the deep learning domain has inspired researchers to exploit neural networks for these tracking applications~\cite{ning2017spatially,sun2019deep,zhai2018deep,zhu2018distractor,guo2017learning,shen2019visual,yu2020deformable}. However, algorithm latency is often compromised for the sake of accuracy when it comes to real-world deployment of neural network driven trackers. For instance, the FCNT tracker~\cite{wang2015visual} achieves a remarkable AUC score of 0.599 post-deployment but is bottlenecked by its latency performance - it only manages to go up to 3 FPS. In a similar vein, the MDNet tracker~\cite{nam2016learning} also achieves an excellent AUC score of 0.678 on the OTB100 dataset, but is let down by its latency performance (it only manages to attain a speed of 1 FPS).  


To overcome this problem, researchers are now shifting their focus to energy and resource-efficient tracking solutions. Embedded systems are fast becoming popular platforms for deploying such real-time neural network-based tracking frameworks. These embedded vision systems not only ensure energy efficiency, but also preserve task accuracy. Case in point, the recently proposed SkyNet~\cite{zhang2020skynet} achieves an impressive IoU score of 0.716 while operating at 25.05 frames per second and 7.26 W power on an Ultra96 embedded FPGA. Embedded systems forsake generality and thus, ensure high task fidelity and efficiency via hardware customization attuned to task needs. As such, embedded computer vision is gaining traction for low-power, real-time vision applications like tracking~\cite{yang2018embedded,zhang2018minitracker,nousi2019embedded}. However, visual data processing comes with a set of complications posed by system constraints. Typical image sensor readout architectures are notoriously power-hungry~\cite{likamwa2013energy,likamwa2016redeye} and this is a huge bottleneck for developing energy-efficient vision algorithms for real-world applications. To address this issue, there has been increasing research efforts geared towards joint optimization of embedded systems and image sensors for reducing overall system power consumption~\cite{zhao2019real,hossain2019deep,palossi201964,hussain2019cmos}. 


Opting for embedded computing environments provides us with the freedom and flexibility to develop and implement highly custom mechanisms targeted towards energy conservation. One such mechanism is based on the notion of region-of-interest (ROI). The ROI is defined by a bounding box enclosing the area of an image frame in which the target object exists. ROI-based energy optimization entails the discarding of pixels outside of the ROI and selectively reading out only the pixels comprising the moving object. The resulting benefits are two-fold: faster bandwidth and improved energy efficiency (owing to selective readout)~\cite{saha2018adaptive,wang2016energy}.
Further, the reduction in pixels also has implications in the post-processing stage, where fewer pixels imply fewer clock cycles in the ISP pipeline and in the end vision task. Fewer pixels also alleviates the computational burden and frees up on-board memory and resources to be used up for other tasks. ROI technology can also potentially increase the frame rate of the camera. However, the key challenge now is to adapt existing tracking algorithms for ROI-capable image sensors. We refer to this new class of tracking algorithms as adaptive subsampling algorithms. 

In this work, we present an extensive study on adaptive subsampling algorithms featuring various object detectors (both classical and machine learning-based) and we evaluate their performance in an adaptive subsampling setup. We also evaluate the deployment of a Kalman filter as an ROI predictor to help improve the subsampling performance of these detectors by correctly predicting the future object trajectories and making decisions accordingly. 

Further, we aim to accelerate these algorithms on FPGAs and study their performance. FPGAs are widely used for low-power and high-latency applications and therefore, are the perfect candidates for implementing energy-efficient adaptive subsampling. Comparing FPGAs to GPUs, we must consider the high power consumption of the latter and the reconfigurability of the former. The obvious choice then becomes FPGA acceleration for our particular use case. In order to implement our algorithms, we leverage Xilinx's Vitis AI tools for mapping our neural network-based subsampling algorithms onto the FPGA. We opt for high level synthesis (HLS) over RTL mapping because of the former's ease of use. Our experimental results show the HLS mapping in no way diminishes the expected performance of our algorithms, that is, we manage to achieve real-time performance without having to resort to RTL mapping.

In this work, we show how we can couple off-the-shelf object detectors with a Kalman filter to jointly perform predictive object tracking and adaptive subsampling. Our paper builds on initial work presented in~\cite{9191146}, and we extend that work by introducing several new types of object detectors to the adaptive subsampling pipeline, and evaluating these algorithms on more comprehensive test datasets. In addition, we also identify a suitable candidate for hardware acceleration, and map the neural network-based approach (Efficient Convolution Operators for Tracking (ECO) plus Kalman filter) onto an FPGA. We also show hardware acceleration results with the YOLO plus Kalman filter-based method for comparative evaluation of the ECO-based approach. Averaged across our two benchmarking datasets, the YOLO-based algorithm achieves an AUC score of 0.2721 while the ECO-based approach results in an AUC score of 0.4020. This, coupled with the fact that the we get higher power savings with the ECO method, makes the ECO+KF framework the ideal backbone of our adaptive subsampling algorithm. As per our power consumption model, the YOLO-driven approach requires approximately 6 W while the ECO method requires only 4 W. Additionally, the ECO+KF method operates at 19.23 FPS, which is nearly real-time. Although the YOLO-based method is faster, it is also less accurate. In terms of accuracy-latency tradeoff, the ECO-driven approach is the clear winner.   

Finally, we would like to highlight the novelty of our work. We have identified a scope for energy optimization in image sensors, and we have shown how predictive visual tracking can be exploited for selective ROI readout for power-hungry vision applications. In order to do this, we have shown how we can harness the predictive capabilities of the Kalman filter in conjunction with a suitable object detector to adaptively subsample image frames and save energy. In our paper we not only demonstrate a new method for mitigating power consumption of the ISP pipeline, but also focus on the hardware-software co-design of an adaptive subsampling system. To our knowledge, this is the first work that demonstrates digital ROI-ing capabilities of the ECO tracker with a Kalman filter on an embedded platform. We have identified the ECO tracker as the candidate having an ideal energy-accuracy tradeoff profile and have mapped it onto an FPGA. Further, we have tested the capabilities of the algorithm on the FPGA by streaming in video data from a webcam.  

\begin{figure}[!t]
\begin{center}
   \includegraphics[width=0.4\textwidth]{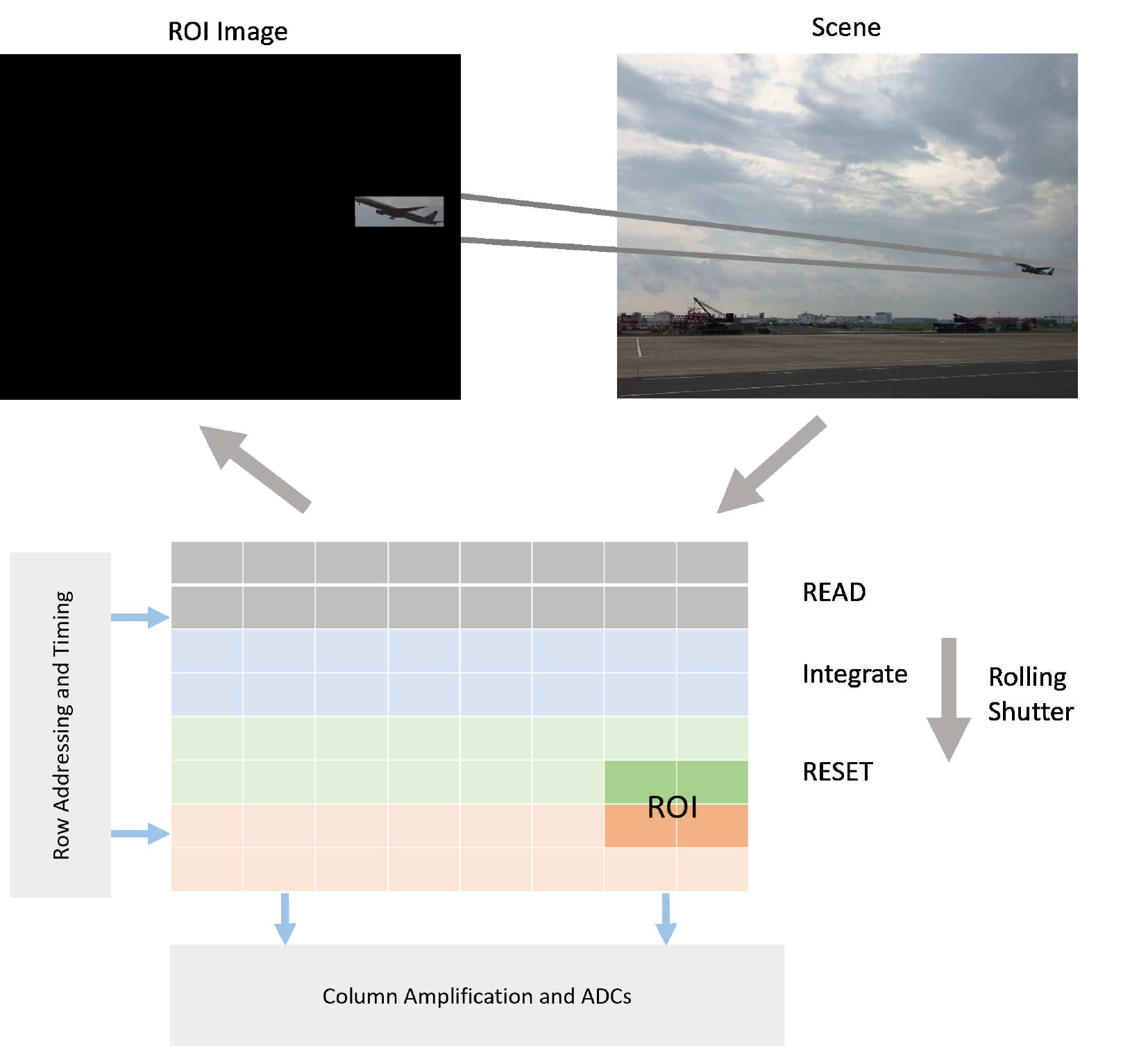}
\end{center}
   \captionsetup{skip=-2mm}
   \caption{How image sensor reads out an ROI image using a programmable rolling shutter. The rolling shutter mechanism helps capture and integrate temporal and spatial data by adapting the readout timing and exposure length of each pixel row. Here, an ROI image has been obtained using the windowing technique - where pixels outside of the object of interest have been switched off. Energy efficiency can also be improved by skipping (skip every other pixel row or column and thereby read out fewer pixels) and binning (group adjacent pixels together and represent them by a single value, thus resulting in fewer pixels to be read out). 
   }
\label{fig:ROI}
\end{figure}

\section{Related Work}
\vspace{-0.2cm}
The readout architectures of modern image sensors are power hungry, and this causes problems for vision tasks like object tracking which require frame-by-frame processing. This is why
energy-efficient object tracking has been studied over the years~\cite{xu2003localized,casares2011adaptive,hsu2012poot,fuemmeler2010energy,yang2015accurate}. Researchers have also been striving to optimize hardware for tracking tasks~\cite{lin2018architectural, han2018low, yamaoka2006image, yamaoka2006multi}. 


In this work, we derive inspiration from other works that have exploited the Kalman filter for tracking~\cite{li2010multiple,black2002multi,marcenaro2002multiple,kim2014data,9191146}, and devise our own method for energy-efficient object tracking using the Kalman filter. In our energy-efficient tracking pipeline, we leverage the notion of region of interest (ROI) and adaptive subsampling. ROI refers to the region of an image frame which encloses the object of interest, and adaptive subsampling refers to the reading out of just the ROIs and the discarding of all pixels not located within the region of interest. ROIs have been exploited for image compression~\cite{lin2006adaptive,belfor1994spatially} in the past. Recent studies have been diving into how ROI can be leveraged for speed benefits~\cite{kodukula2021rhythmic, tizon2011roi, lu2017hybrid}. Other works have discussed how we can power-gate image sensor resolution for high energy savings~\cite{buckler2017reconfiguring}. 


Existing tracker paradigms can be boiled down to two main classes: Discriminative correlation filter (DCF)-based (such as ECO~\cite{danelljan2017eco}) and Siamese network-based
trackers (such as SiamRPN~\cite{li2018high}, ATOM~\cite{danelljan2019atom},DimP~\cite{bhat2019learning}). Although DCF-based approaches are highly efficient, they are normally not on par with the Siamese network-driven methods. ECO has managed to bridge the gap to some extent. ECO uses network-based feature extraction but still downsizes by employing factorized convolution operations. ATOM and DiMP, although state-of-the-art, are large sized networks and are not light-weight. That is why we opted for accelerating the ECO-based method on an FPGA. In addition, we also accelerated the YOLO and Tiny-YOLO-based algorithms. However, the ECO shows better performance than the other two in terms of the benchmarking datasets we used. 


State-of-the-art object tracking utilizes deep neural networks~\cite{son2017multi,redmon2016you, fan2017sanet, nam2016learning, song2018vital, wang2015visual}. However, none of these works deal with adaptive image subsampling and are not concerned with the massive power consumption associated with continuous image sensing. In addition, deep neural networks are notoriously compute-heavy and often have high-latency performance even in hardware. In this study, we have selected the most computationally efficient neural network-based object detectors for our adaptive subsampling pipeline and have further accelerated them via FPGA implementation. Note that in our pipeline, frames are read out in their entirety only when the object detector is activated. At all other instances, the Kalman filter is employed for making ROI predictions which further improves the latency and ensures higher energy savings. 

In~\cite{yao2018semantics}, a tracking algorithm has been proposed that features semantic awareness for category-oriented object proposals. In the proposed tracking pipeline, semantic information has been leveraged to localize the object of interest using a scale adaptation method. In~\cite{han2018low}, a low-power neural processor has been designed for mobile devices to achieve real-time object or ROI tracking performance. In~\cite{casares2011adaptive}, energy consumption and processing time are reduced by a feedback method wherein the tracking information of the previous frame is used to compress the search area of the consecutive frame’s ROI, thus saving time. These works have some similarities to our work in terms of efficient single-object video tracking, but our paper focuses on efficiently reading out only relevant visual data via the ROI of a camera and exploring the energy-accuracy tradeoff for doing so, and by developing an FPGA implementation for our predictive tracking algorithm in this journal paper.  We feel this distinguishes our work significantly from the previous papers in the field. 

There exist a couple of works in the literature that resonate very closely to ours - objectness-based subsampling~\cite{mohan2019adaptive} and predictive visual tracking~\cite{li2021predictive}. In the former, the ROI is localized by a feature called objectness. A heat map of probable locations is computed, and Otsu's threshold is used to determine an ROI and corresponding subsampling mask. However, this algorithm defaults back to sampling full frames whenever there is significant motion. In our paper, we adaptively adjust the ROI using both classical means and deep learning methods, and this equips us to handle fast moving objects. In~\cite{li2021predictive}, a Kalman filter-based predictive tracking method similar to ours is proposed. The authors show how tracking via forecasting can be exploited to attain faster latency, thereby bridging the gap between theory and real-world implementation. The computationally inexpensive Kalman filter is used as a forecaster, and frames are skipped whenever the tracker is slower than the world frame rate.  
Therefore, the focus in this paper is purely on latency-aware tracking. In contrast, we emphasize on energy savings using adaptive subsampling. 

\section{Approach}
\label{sec:M}

\begin{figure}[!t]
\begin{center}
   \includegraphics[width=0.48\textwidth]{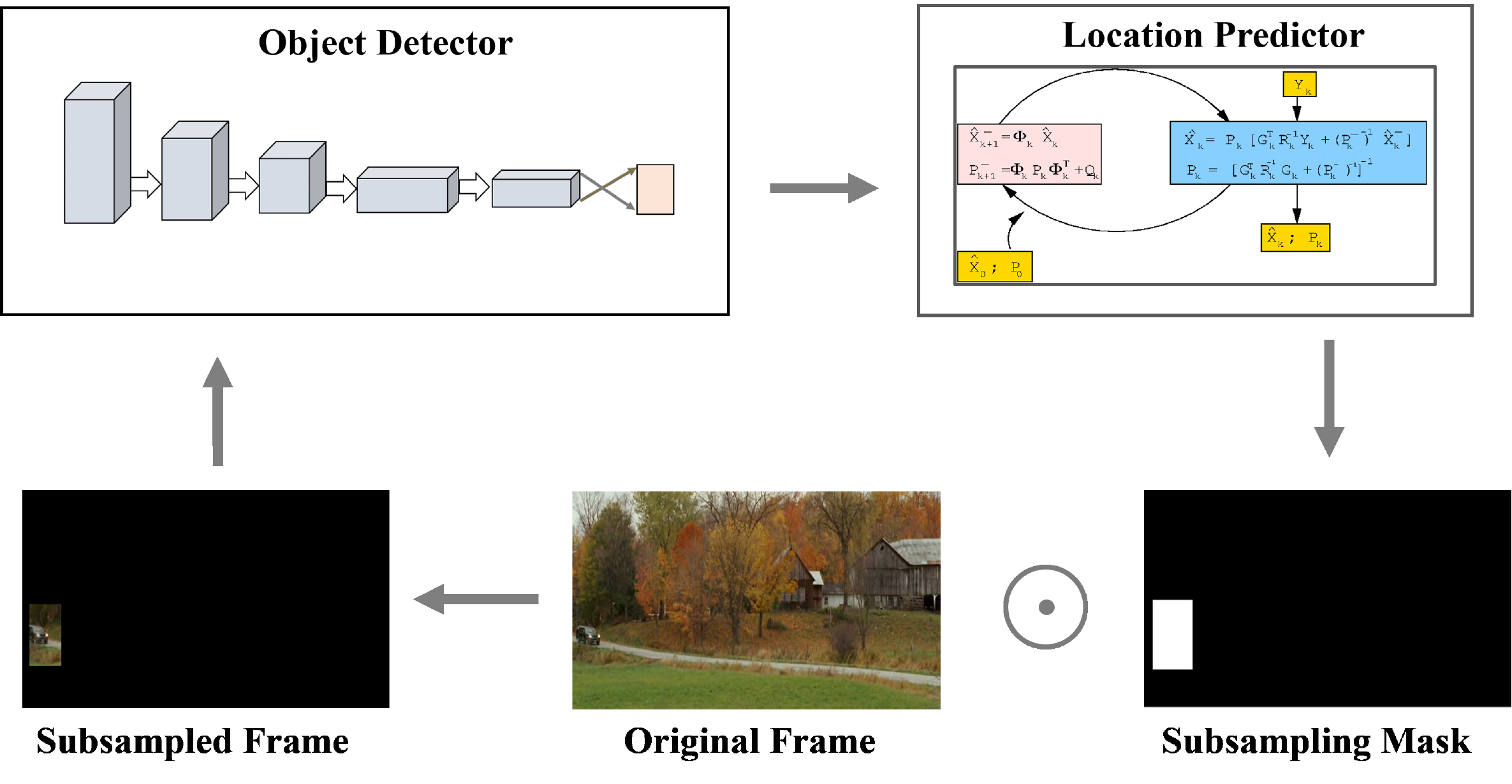}
\end{center}
   \captionsetup{skip=-2mm}
   \caption{An object detector identifies the ROI and this information is fed as external measurements during the Kalman filter's update phase. The Kalman filter then makes predictions while the non-keyframes are read out from the image sensor.
   }

   

\label{fig:block_diagram}
\end{figure}
In this section, we give a detailed description of the video subsampling and ROI prediction problem, and show how we can go about solving it. 

\subsection{Video Subsampling and ROI Prediction} 

Configuring the image sensor on-the-fly such that the pixels outside the ROI are not read out is referred to as adaptive subsampling~\cite{9191146}. Adaptive subsampling when applied in the context of sequential frame capture is referred to as video subsampling. In video subsampling, ROI is read out for one image frame $I(x,y,t)$, and this information is utilized to determine where the object might move in the following frame $I(x,y,t+1)$ at the next time step. To preserve energy, sensor pixels are switched off by employing an algorithmically determined ROI mask $\hat{M}(x,y,t)$. The mask $\hat{M}(x,y,t)$ is of the same shape as the image frame, and it is all '1's inside of the region of interest and all '0's outside of the ROI. The scene at time step $t+1$ undergoes a Hadamard product operation with the ROI mask determined at time step $t$ and the resulting image is referred to as a subsampled image. That is, subsampled image $I_{subsampled}(x,y,t+1) = \hat{M}(x,y,t) \odot I(x,y,t+1)$. Thus, ROI localization is achieved via predictive tracking. It is possible to create ROIs with various geometric shapes like circles/ellipses or even arbitrary shapes. In this work, all our ROIs are rectangular since that is most commonly supported by existing image sensor hardware. 

In our video subsampling pipeline, we use a Kalman filter (KF)~\cite{kalman1960new} for making these predictions. In the update phase of the Kalman filter, where it requires  external measurements to updates its state estimate vector, we invoke an object detector $D(\cdot)$ which operates on a fully sampled image frame to localize the ROI. The detector output $D(I(x,y,t)) = b_t$ is a vector containing the bounding box coordinates of the current frame, and this vector $b$ is used as the external measurements in the update phase of the Kalman filter. In the prediction phase, the Kalman filter solely relies on its own state space matrix and perceptual capabilities to identify the location of the target object. At first, a state prediction step is executed by the filter wherein the current state space $x_{t}$ and state transition matrix $A$ are used to predict $x_{pred}$:
\begin{equation}
    x_{pred} = Ax_{t}.
\end{equation}
Note that, the state space $x_t$ denotes the bounding box coordinates at time step $t$. 
Another crucial step in the prediction phase of the filter is to make a prediction on the covariance matrix as follows:
\begin{equation}
    P_{pred} = AP_tA^T+Q,
\end{equation}
where, $P_t$ is the covariance matrix at time step $t$ and $P_{pred}$ is the covariance matrix prediction for time step $t+1$. Here, the term Q represents the process noise covariance. 

In the update phase of the Kalman filter, the detector output $D(\cdot) = b_t$ at time step $t$ is utilized as follows:
\begin{equation}
    y_t = b_t - Hx_{pred}
\end{equation}
where, H is the observation model and $x_{pred}$ was the last Kalman filter prediction made at time step $t-1$. The time indices are critical here; when the object detector $D(\cdot)$ is invoked, the Kalman filter remains inoperational and waits for the correctional bounding box information from the detector. Hence, the last $x_{pred}$ would be the one obtained at time step $t-1$. At time step $t$, the object detector takes over and the Kalman filter shifts to the update phase before the prediction phase again kicks into motion from the next time step $t+1$. 

After computing $y_t$, the innovation covariance $S$ is obtained as follows:
\begin{equation}
    S = HP_{pred}H^T + R,
\end{equation}
where, R is the covariance of the observation noise. Subsequently, the optimal Kalman gain for time step $t$ can be computed in the following way:
\begin{equation}
    K = P_{pred}H^TS^{-1}.
\end{equation}
Once we have the gain, the state space is updated to get the state space at time step $t+1$:
\begin{equation}
    x_{t+1} = x_{pred} + Ky_t.
\end{equation}

This $x_{t+1}$ term obtained from the filter is used to define the sensor mask $\hat{M}(x,y,t+1)$ for time step $t+1$. The bounding box coordinates defined by $x_{t+1}$ are used to identify where the mask would have '0's and where it would have '1's. 

And finally, the covariance matrix is updated as follows:
\begin{equation}
    P_{t+1} = (1-KH)P_{pred}.
\end{equation}

Granted, extended Kalman filters, particle filters, etc. might prove to be more accurate and better-suited to our goal of ROI prediction. However, we must also take into account the accompanying resource utilization and on-board clock cycles requirement associated with FPGA acceleration. The Kalman filter is an FPGA compatible and lightweight tracker and, therefore, is ideal for addressing problems relating to latency and computational efficiency. This is why we have chosen the Kalman filter as our prime candidate for ROI prediction. 

Notice how the interval between the update and prediction steps of the Kalman filter is programmable. We can delay or prolong the activity of the object detector for as long as we like. We refer to the tracking intervals during which prediction takes place as keyframing intervals. For the Kalman filter-based methods, the keyframing interval would be the interim in which the Kalman filter is made to predict future ROIs without any input from the object detector. If the keyframing interval is defined by $k$ number of frames, then the prediction phase of the Kalman filter will last from time step $t$ till $t+k$. Subsequently, the previously dormant $D(\cdot)$ operator will be activated at time step $t+k+1$, and a fully sampled image frame $I(x,y,t+k+1)$ will be fed through it as follows:
\begin{equation}
    D(I(x,y,t+k+1)) = b_{t+k+1}.
\end{equation}
The frames which are readout in their entirety are referred to as key frames and these are the image frames that are sent to the object detector $D(\cdot)$ for processing. In this instance, $I(x,y,t+k+1)$ is a keyframe. 

At the same time step $t$, the Kalman filter will use the correctional signal from the detector $b_{t+k+1}$, and undergo the update phase. Thereafter, the prediction phase of the filter will again kick into gear from time step $t+k+2$ till $(t+k+2)+k$ - and in the process it will assist in generating subsampled images $I_{subsampled}(t+k+2)$ through to $I_{subsampled}((t+k+2)+k)$. Thus, the process will continue till the last frame. 

The keyframing interval is a primary way to tradeoff between detection and predictive capability of the Kalman filter. The longer the interval, the higher the optimization of the energy and computational efficiency. However, the worse will be the tracking precision as the Kalman filter provides only a simplistic object trajectory model. Hence, it is a matter of the demands of the end task how the precision and energy requirement are to be prioritized. The keyframing interval can be tuned accordingly. 


\subsection{Object Detectors} In our search for a FPGA-compatible object detector that forms the backbone of our adaptive subsampling algorithm, we have studied various off-the-shelf detectors ranging from classical to deep learning-based. 

\textit{Mean Shift Tracking (MS).} 
The mean shift algorithm leverages the color histogram of an image to keep track of the ROI as a cluster of color histogram values~\cite{fukunaga1975estimation}. Note that the notion of coupling the Kalman filter with a mean shift tracker is not a novel concept as evidenced by~\cite{ren2012mean}. However, our algorithm is different in that we set it up as an adaptive subsampling algorithm as in~\cite{9191146}. Also, we conduct a comparative study with other FPGA compatible algorithms and present better methods that outperform this technique.

\textit{Kernelized Correlation Filter (KCF).} 
KCF iteratively estimates current object location based on the past ROI localization~\cite{henriques2014high}. Visual features are extracted from the FFT-based ROI selected by KCF, and these are used to update the appearance model as well as the correlation filter information. For predictive tracking, we couple this KCF-based tracker with a Kalman filter.


\textit{Distractor-Aware Tracker (DAT).} The DAT tracker performs online object tracking based on color representations of images~\cite{possegger15a}. DAT incorporates a discriminative object model which preemptively identifies potential ``distracting" regions and steers away from them to remain on the correct trajectory. This tracker is also coupled with the Kalman filter as described earlier for predictive tracking.

\textit{YOLO and Tiny-YOLO CNN.} The YOLO neural network architecture proposed by Redmon et al.~\cite{redmon2016you} is a real-time object detector and is a good candidate for tracking applications. We evaluate both the pre-trained YOLO and the pre-trained tiny-YOLO convolutional neural networks in the measurement phase of the Kalman filter. These networks can optimize the size of their ROI by changing the width and height of the bounding box frame to frame. Additionally, the tiny YOLO's lightweight architecture and its end-to-end optimization framework make it an attractive object detector for mobile and real-time applications. 




\textit{Accurate Tracking by Overlap Maximization (ATOM).} ATOM utilizes high-level target information for offline learning and then employs a dedicated classification component for online learning of object trajectories~\cite{danelljan2019atom}. 

\textit{Learning Discriminative Model Prediction for Tracking (DiMP).} DiMP specializes in leveraging both foreground and background information for ROI estimation~\cite{bhat2019learning}. 

While both ATOM and DiMP demonstrate remarkable tracking capability, we show in our experimental results that they do not sustain their performance for adaptively subsampled images. To adapt these two methods for operating in an adaptive subsampling setup, we modify the algorithms and allow the output ROI of one image frame $D(I(x,y,t)) = b_t$ dictate the ROI sensor mask of the next incoming frame $I(x,y,t+1)$. That is to say, we treat the output of the first frame as a prediction of where the object will appear in the next frame. Using this location information $b_t$, we generate sensor mask $\hat{M}(x,y,t)$ and subsample the next incoming frame as follows:
\begin{equation}
    I_{subsampled}(x,y,t+1) = \hat{M}(x,y,t) \cdot I(x,y,t+1).
\end{equation}
Afterwards, this subsampled image $I_{subsampled}(x,y,t+1)$ is what gets fed to the network. That is, $D(\cdot)$ now operates on a subsampled images as follows:
\begin{equation}
    D(I_{subsampled}(x,y,t+1)) = b_{t+1}.
\end{equation}
The process continues until the keyframing interval ends, which is when the $D(\cdot)$ operator once again receives a fully sampled image frame so that it can correct its trajectory if it has veered off course. This enables performance analysis of the ATOM and DiMP methods in the context of adaptive subsampling. Further, we also study what bearing the Kalman filter has on the performance when we incorporate it into the ATOM and DiMP based adaptive subsampling algorithms. The Kalman filter makes the predictions and these predictions are used to subsample the pixels outside of the ROI - exactly like we discussed in the previous subsection. In this variation of the algorithm, the ATOM and DiMP are relegated to the sole purpose of feeding external measurements to the Kalman filter after certain intervals such that the Kalman filter is able to update and correct its trajectory. We also set up the other object detectors to operate in similar fashion, where the output ROI locations of these algorithms are leveraged by the Kalman filter for course correction and ROI prediction. 

\textit{Efficient Convolution Operators for Tracking (ECO).} 
ECO-based tracking was proposed as a solution for the endlessly increasing complexity of discriminative correlation filter-based trackers~\cite{danelljan2017eco}. ECO incorporates a factorized convolution operator for reducing the number of filter parameters, a generative model for better characterizing the input data samples and a model update strategy which updates the model parameters after every few frames. We reformulate the aforementioned model update strategy to introduce our Kalman filter in the pipeline for predictive tracking. 

\begin{figure*}
     \centering
     \begin{subfigure}[b]{0.24\textwidth}
         \centering
         \includegraphics[width=\textwidth]{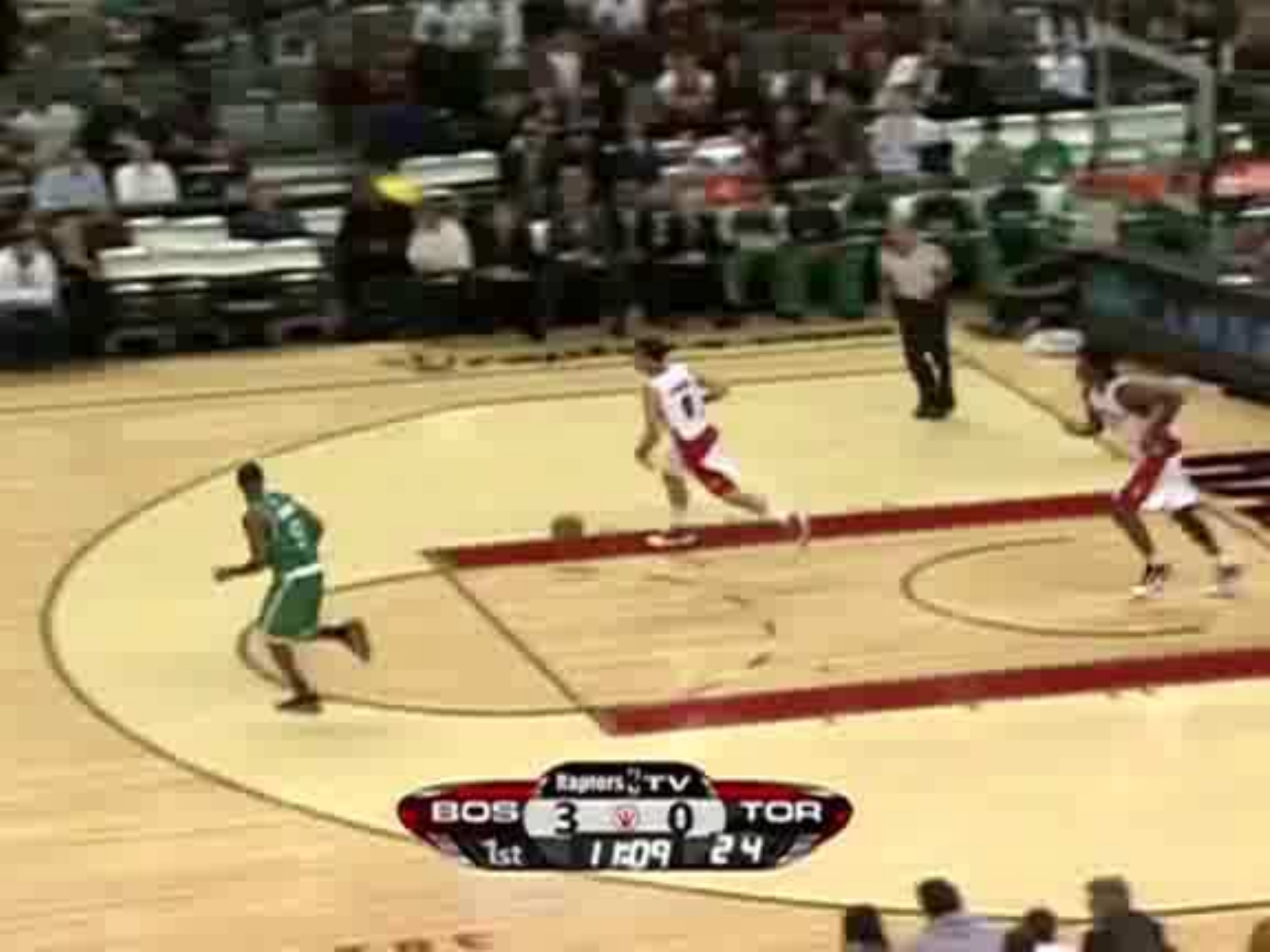}
         \caption{Fully Sampled}
         \label{fig:fully_sampled}
     \end{subfigure}
     \begin{subfigure}[b]{0.24\textwidth}
         \centering
         \includegraphics[width=\textwidth]{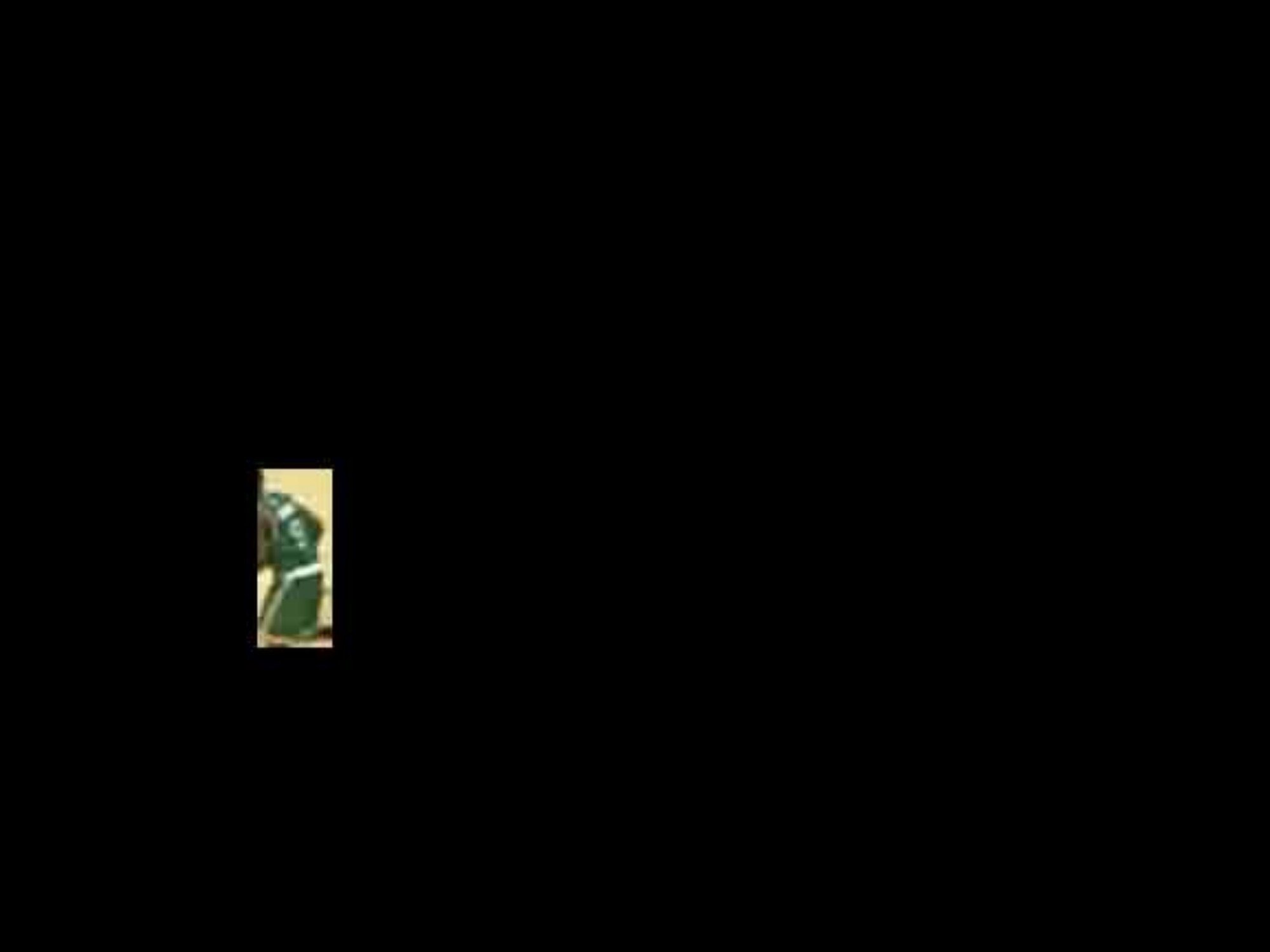}
         \caption{Ground-truth Sampling}
         \label{fig:subsampled}
     \end{subfigure}
     \begin{subfigure}[b]{0.24\textwidth}
         \centering
         \includegraphics[width=\textwidth]{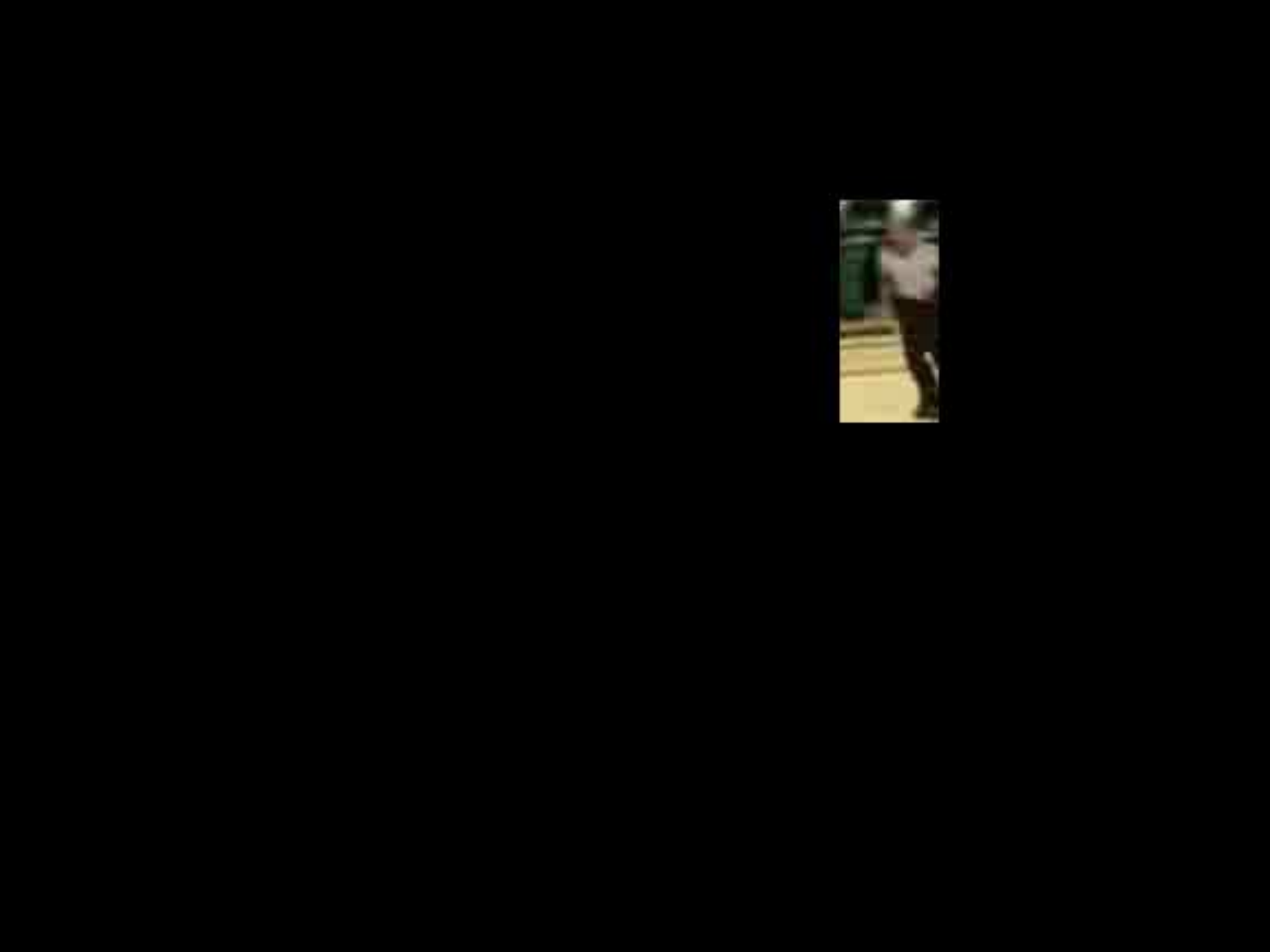}
         \caption{YOLO+KF (IoU=0.0)}
         \label{fig:yolo_kf_sampled}
     \end{subfigure}
     \begin{subfigure}[b]{0.24\textwidth}
         \centering
         \includegraphics[width=\textwidth]{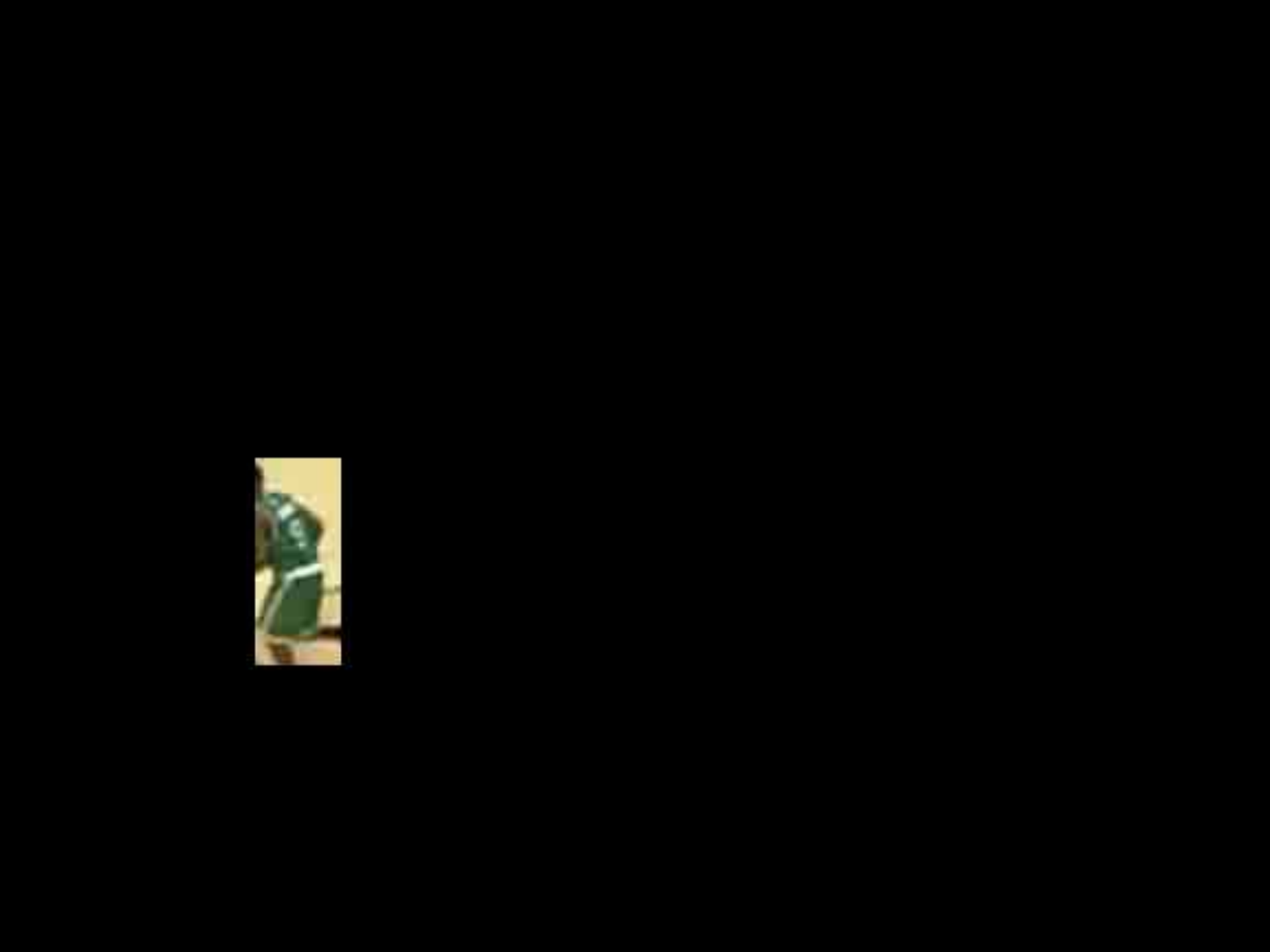}
         \caption{ECO+KF (IoU=0.76)}
         \label{fig:eco_kf_sampled}
     \end{subfigure}
     \begin{subfigure}[b]{0.24\textwidth}
         \centering
         \includegraphics[width=\textwidth]{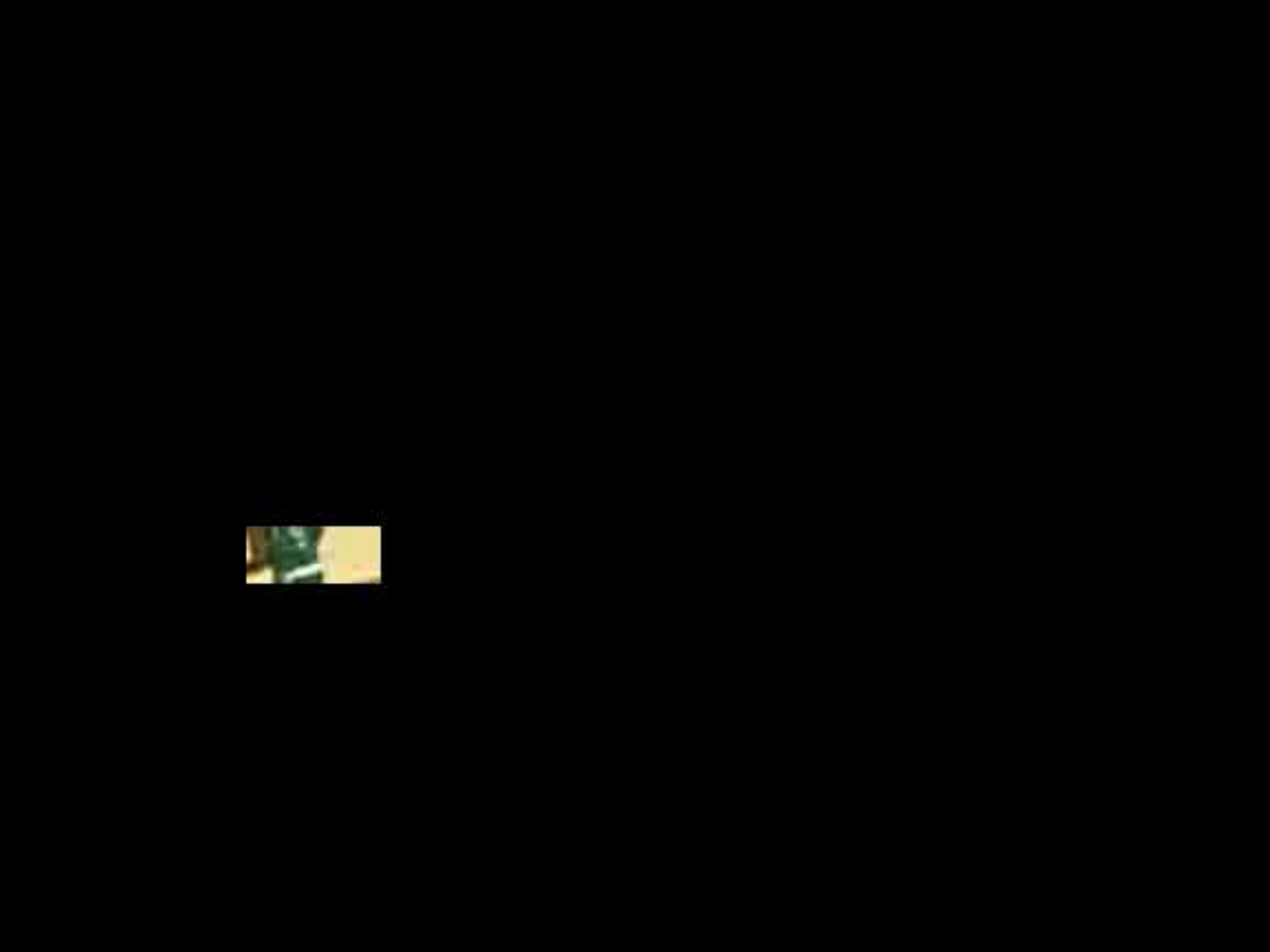}
         \caption{ATOM (IoU=0.26)}
         \label{fig:atom_sampled}
     \end{subfigure}
     \begin{subfigure}[b]{0.24\textwidth}
         \centering
         \includegraphics[width=\textwidth]{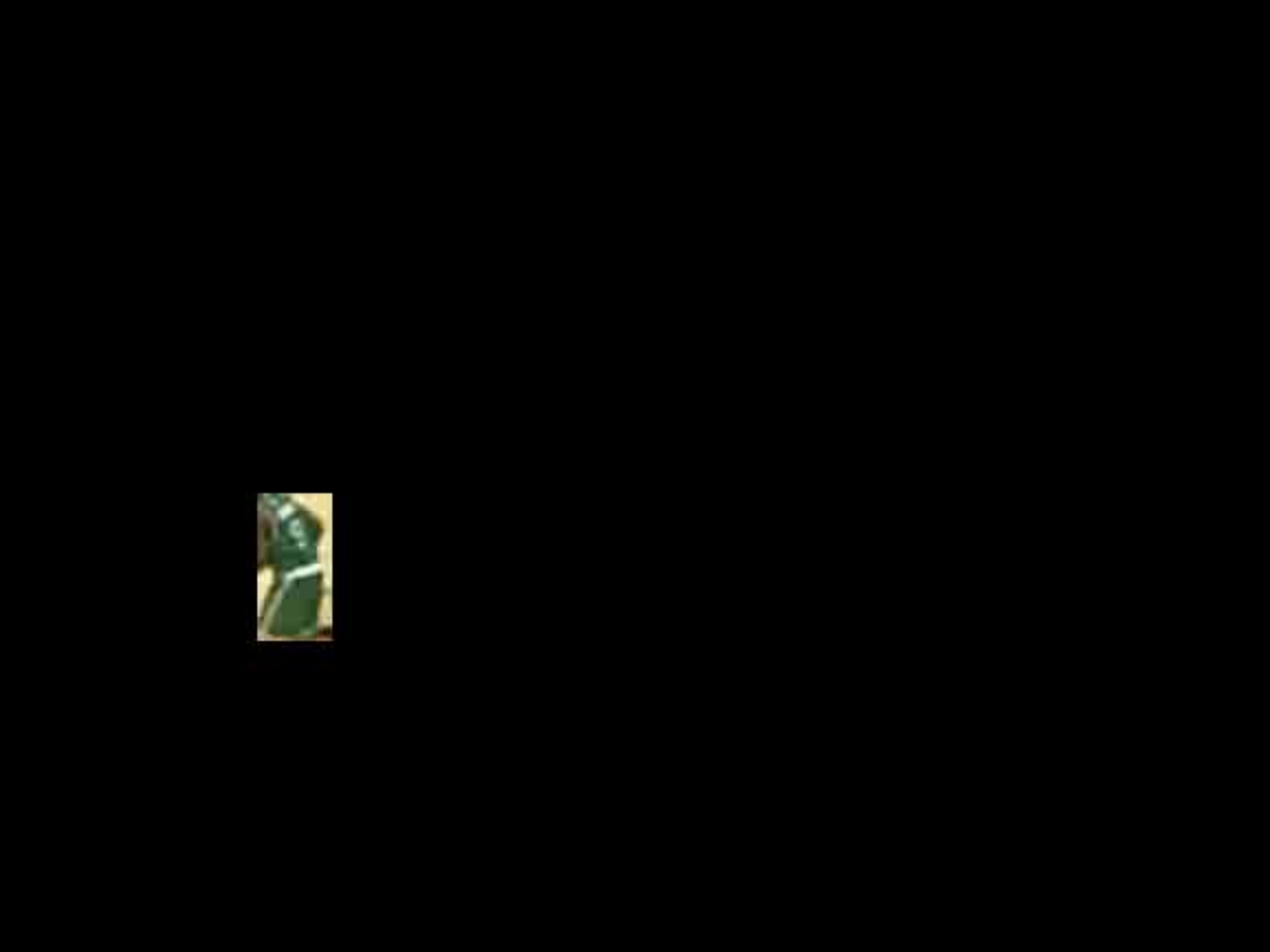}
         \caption{ATOM+KF (IoU=0.83)}
         \label{fig:atom_kf_sampled}
     \end{subfigure}
     \begin{subfigure}[b]{0.24\textwidth}
         \centering
         \includegraphics[width=\textwidth]{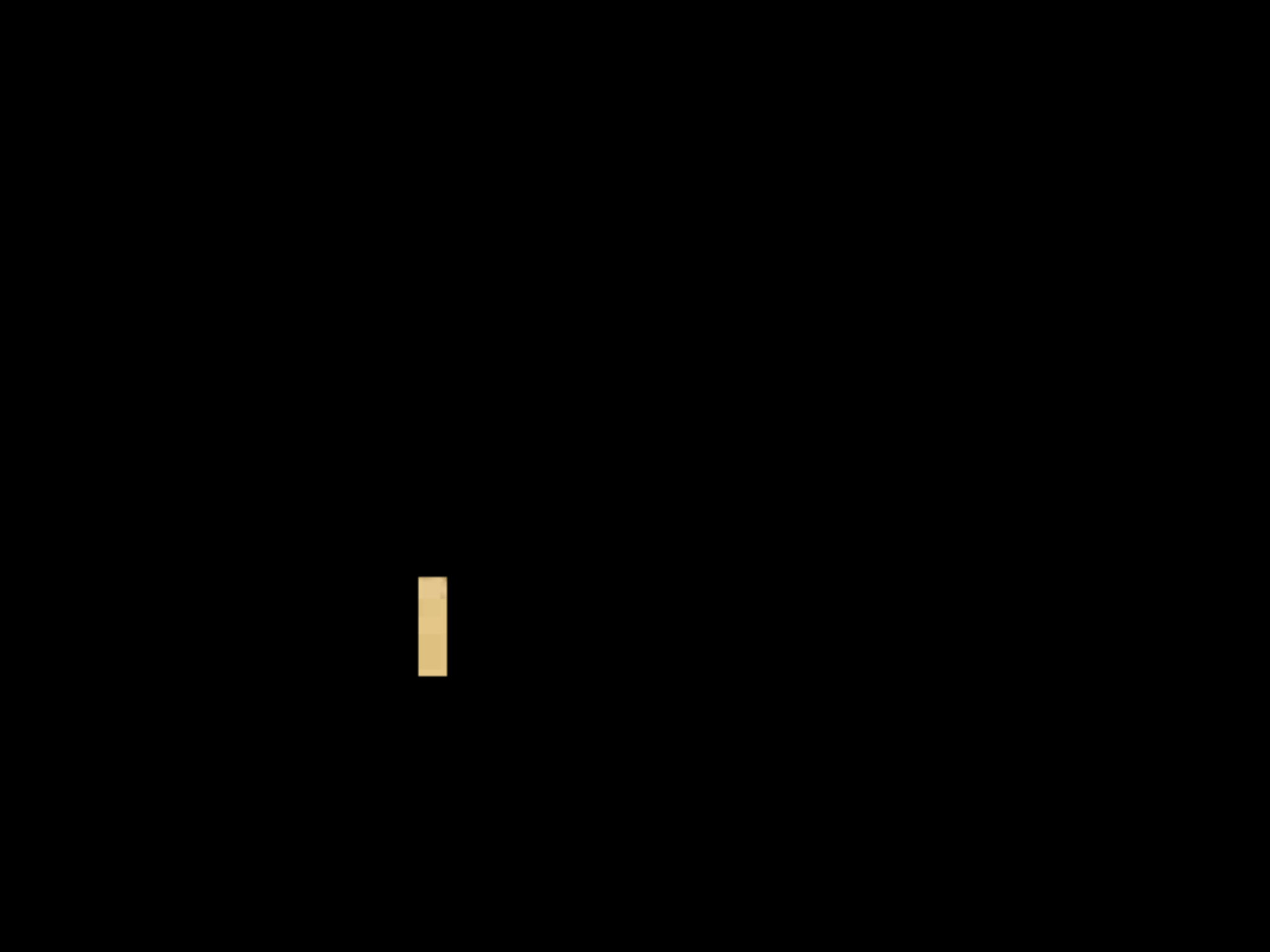}
         \caption{DiMP (IoU=0.0)}
         \label{fig:dimp_sampled}
         \end{subfigure}
     \begin{subfigure}[b]{0.24\textwidth}
         \centering
         \includegraphics[width=\textwidth]{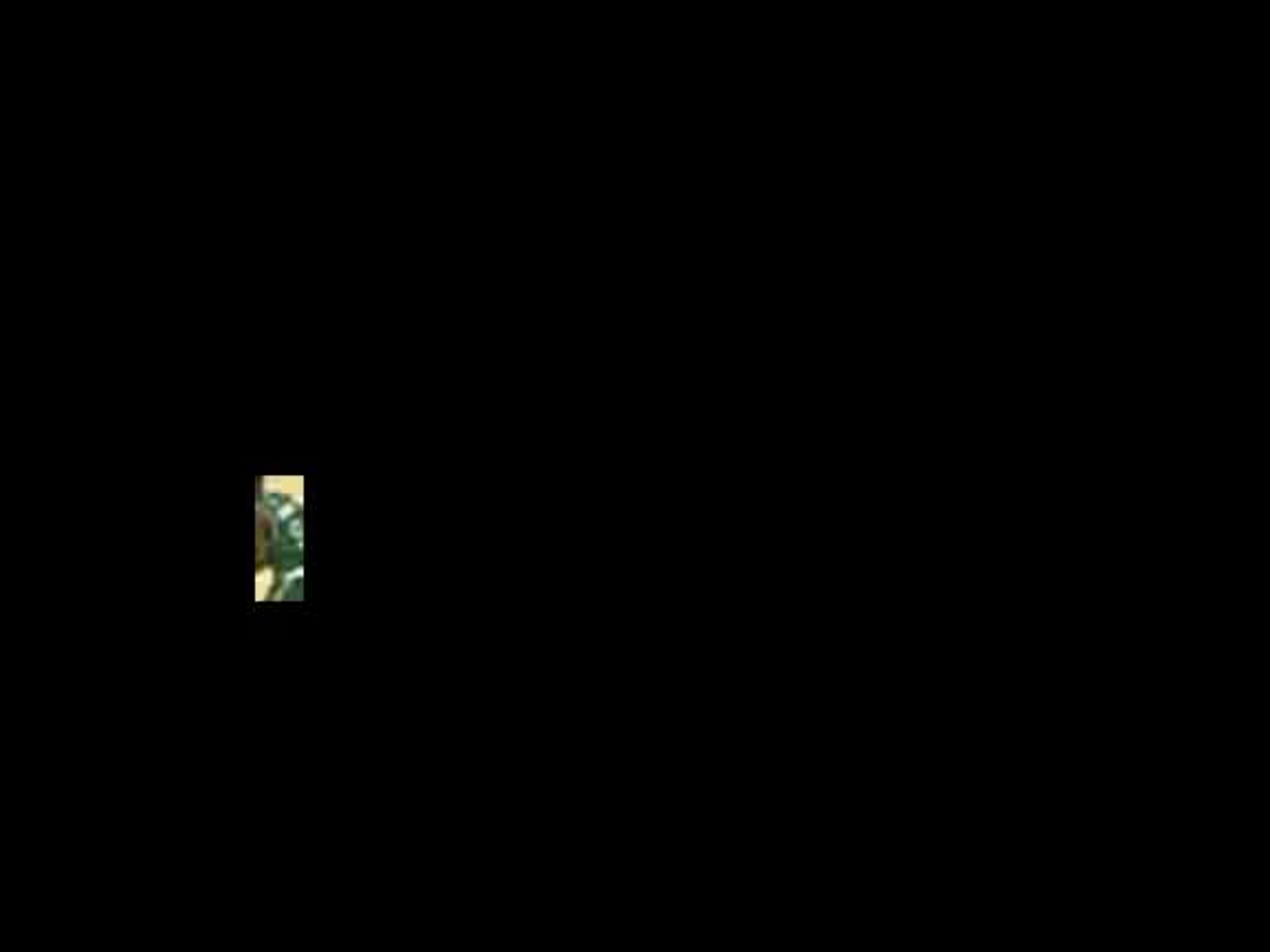}
         \caption{DiMP+KF (IoU=0.46)}
         \label{fig:dimp_kf_sampled}
     \end{subfigure}
     \begin{subfigure}[b]{0.24\textwidth}
         \centering
         \includegraphics[width=\textwidth]{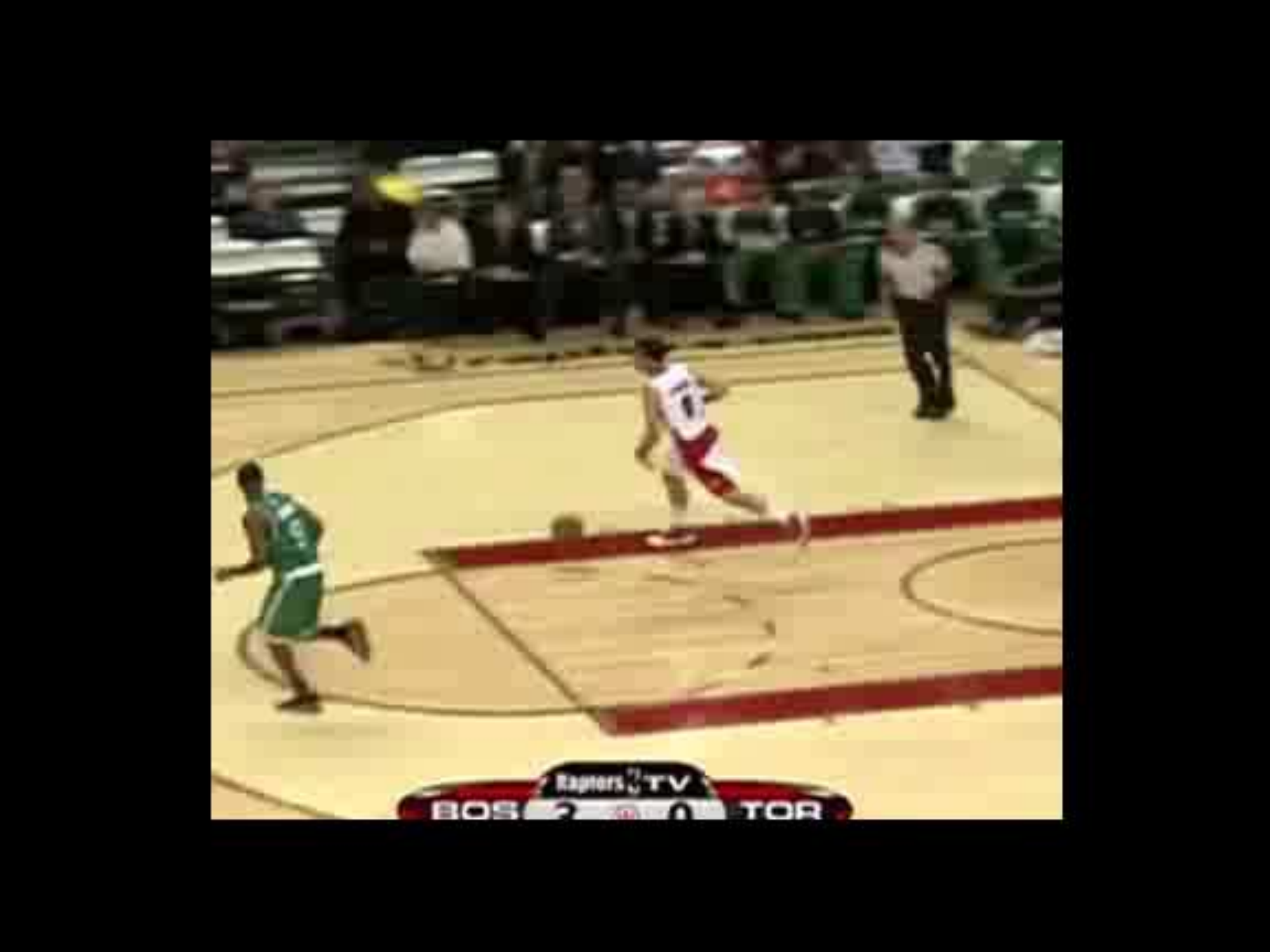}
         \caption{Tiny-YOLO+KF (IoU=0.02)}
         \label{fig:tinyyolo_kf_sampled}
     \end{subfigure}
     \begin{subfigure}[b]{0.24\textwidth}
         \centering
         \includegraphics[width=\textwidth]{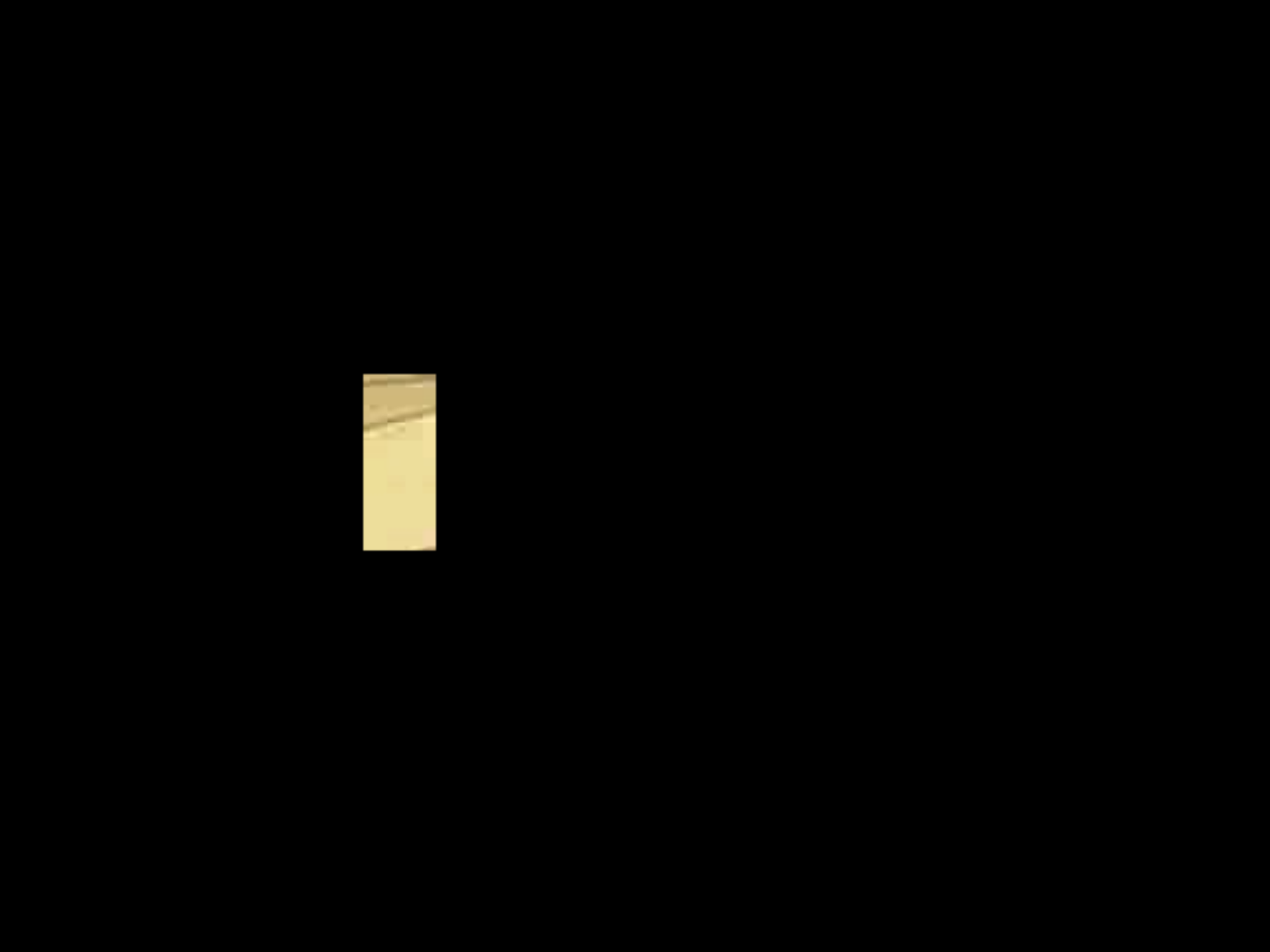}
         \caption{KCF+KF (IoU=0.0)}
         \label{fig:kcf_kf_sampled}
     \end{subfigure}
     \begin{subfigure}[b]{0.24\textwidth}
         \centering
         \includegraphics[width=\textwidth]{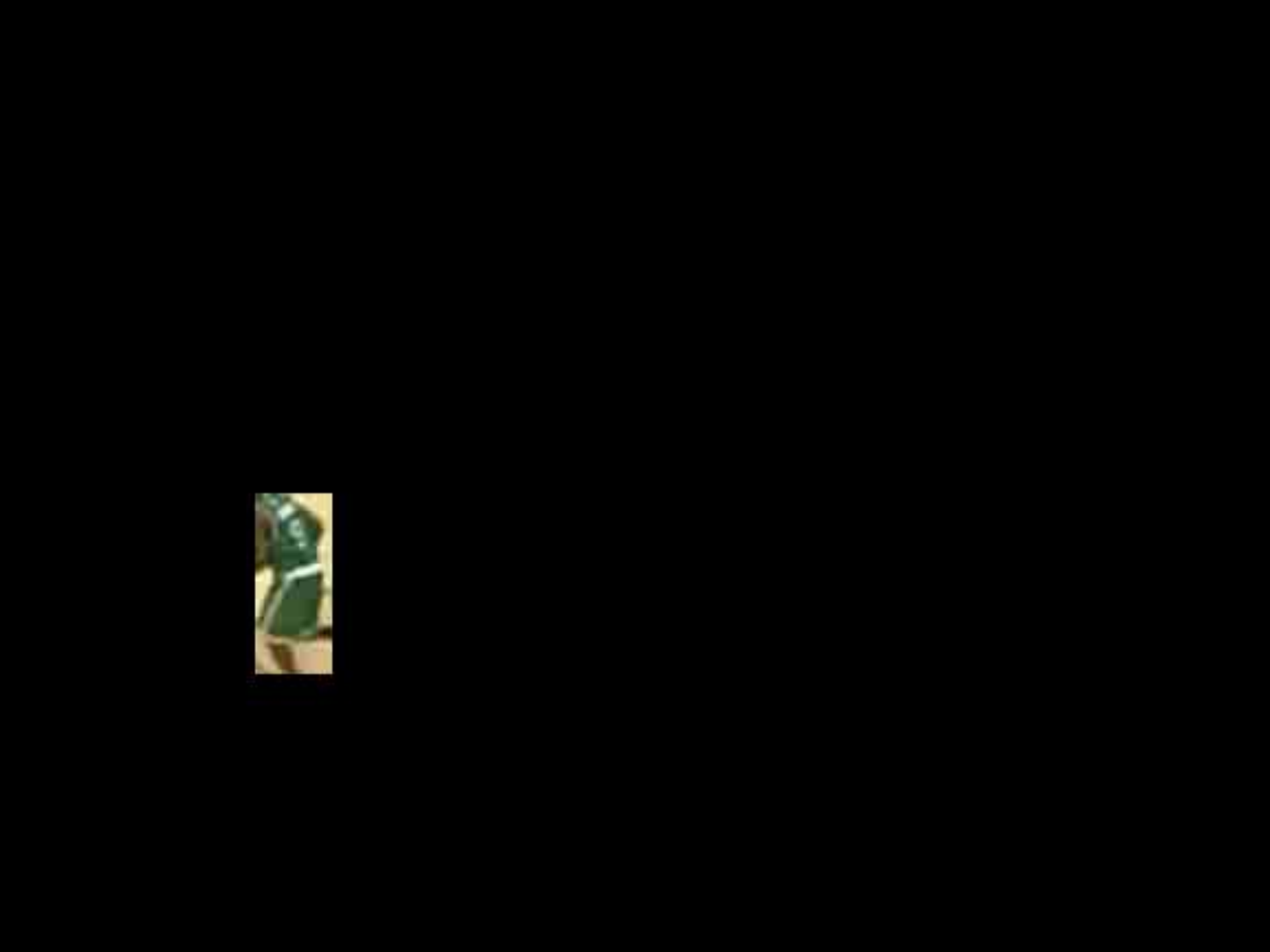}
         \caption{MS+KF (IoU=0.74)}
         \label{fig:ms_kf_sampled}
     \end{subfigure}
     \begin{subfigure}[b]{0.24\textwidth}
         \centering
         \includegraphics[width=\textwidth]{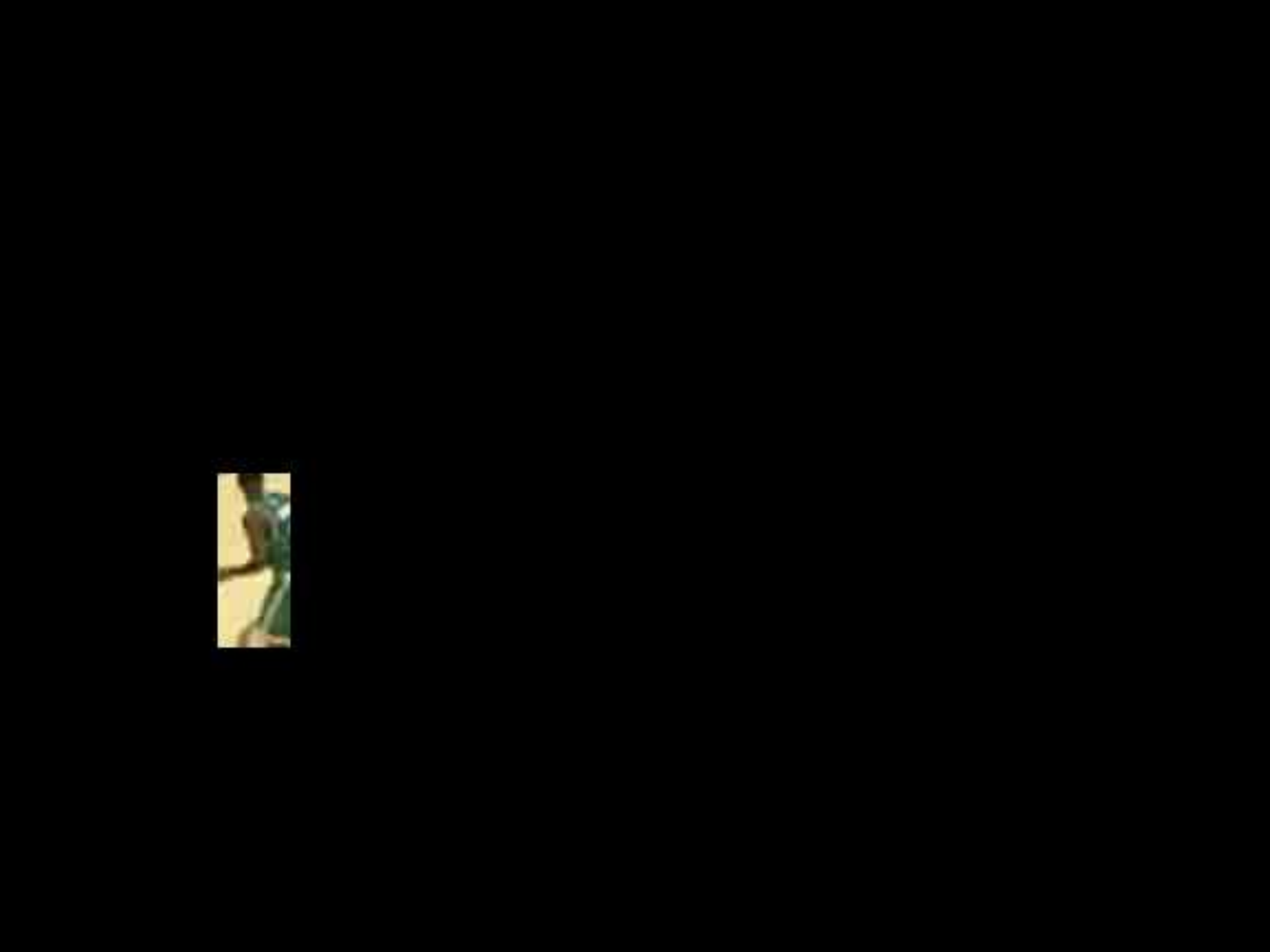}
         \caption{DAT+KF (IoU=0.30)}
         \label{fig:dat_kf_sampled}
     \end{subfigure}
        \caption{Object tracking and subsampling with the various methods. We select the same frame in a video sequence and display the sensor mask generated subsampled image obtained with our selected approaches. The frame generated by the ECO tracker is visually very close to the ground truth and indicates a good mAP score. Further, the ECO tracker has generated a compact bounding box around the object of interest - indicating that only a few of the pixels in the frame will remain activated during frame readout, thus reducing the power consumption.}
        \label{fig:comparison}
\end{figure*}

\section{Algorithm Implementation and Results}
We conduct a set of extensive experiments to evaluate the performance of our adaptive subsampling algorithms in software. These experiments reveal the best candidates for hardware acceleration efforts. In addition, we will show corresponding hardware results in \ref{sec:hardware}, e.g. algorithm latency in hardware and associated resource utilization. 

\indent \textbf{Datasets.} We evaluated our joint adaptive subsampling and tracking algorithms on two benchmarking datasets- the OTB100~\cite{WuLimYang13} and the LaSOT~\cite{fan2019lasot}. The OTB100 dataset is comprised of 100 test video sequences of varying difficulty. Example tasks include tracking a coupon across a table (easy), a basketball player in a sea of similarly attired sportspeople (medium), and a musician on stage in low-light conditions (hard). On the other hand, the LaSOT dataset constitutes a training set and a test set and has a grand total of 1400 videos. Since we employ pre-trained networks for all of our methods, we only use the test dataset from LaSOT which comprises of 280 video sequences to evaluate our algorithms. Example videos from this dataset include a car moving in a low-traffic road in daylight (easy for a tracking task), an airplane zooming in towards the camera from some distance (medium), and a small drone being flown in random patterns in a park featuring vehicles and other distractors (hard). The frame rates of all videos in our test datasets are 30 FPS.




\indent \textbf{Metrics.} Tracking performance of our algorithms has been evaluated in terms of mean average precision (mAP) and area under the curve (auc) scores. We compute the mAP by counting the number of frames wherein the algorithm prediction and ground truth bounding box have an intersection over union (IoU) greater than some pre-determined threshold. If we perform a sweep over the threshold and plot the corresponding mAPs, we obtain a performance curve referred to as a success plot. The area under this curve is the AUC score, and a larger auc score indicates better tracking performance.

\begin{figure*}[!t]
   \includegraphics[width=0.85\linewidth]{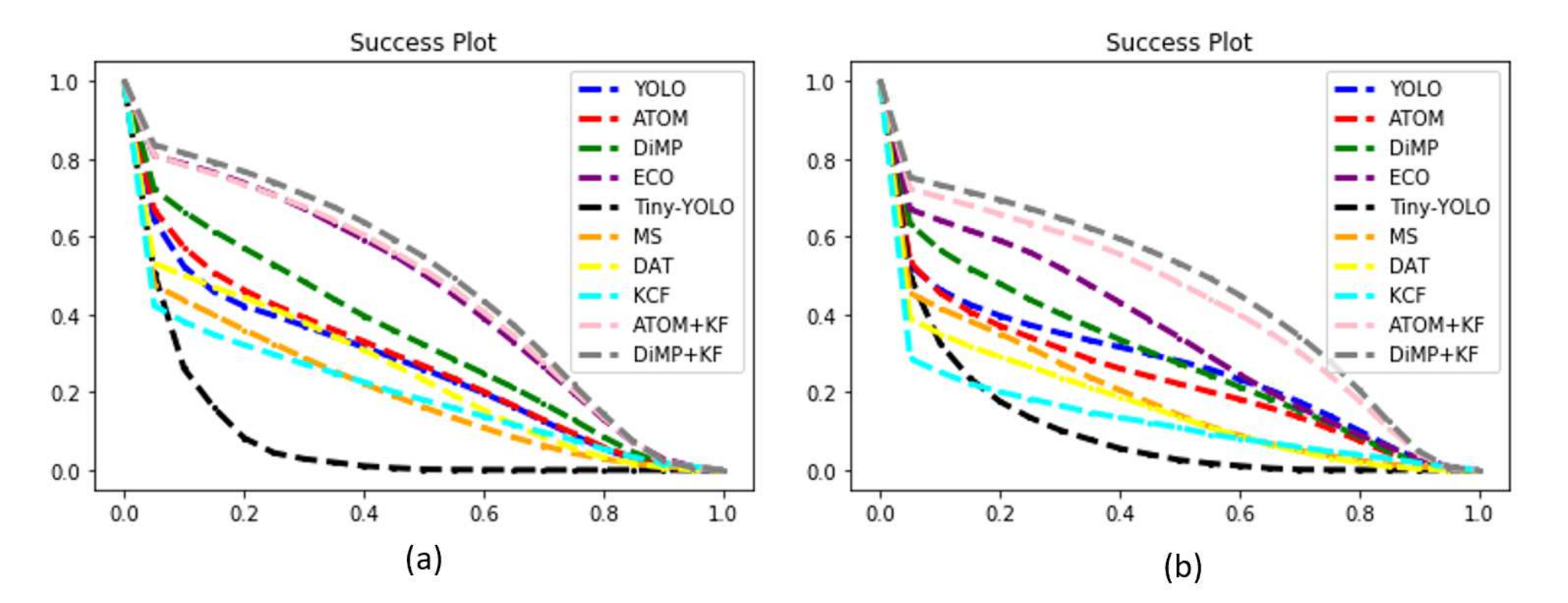}
   \captionsetup{skip=-2mm}
   \caption{Mean average precision results at different IoU thresholds (success plots). We have swept the IoU thresholds from 0 to 1 with a step size of 0.05 for all the adaptive subsampling algorithms on the (a) OTB100 and (b) LaSOT datasets. It is evident that the methods degrade in performance as the constraint on the IoU threshold is increased.}
\label{fig:auc}
\end{figure*}


\indent \textbf{Results.} In Table~\ref{tab:main_results}, we report the AUC scores for the adaptive subsampling algorithms. The DiMP+KF method outperforms the other methods on both the datasets - on OTB100 it attains an AUC score of 0.4817 and on LaSOT it achieves a score of 0.4702. However, the ATOM+KF and ECO+KF can be considered close contenders. As we will later show that the ECO+KF method is best suited for FPGA acceleration and this will be our selected candidate, notice its AUC score on the OTB100 dataset - 0.4568. This is comparable to what the ATOM+KF achieves (0.4625) and is quite close to what we get with DiMP+KF (0.4817). On LaSOT, the ECO+KF demonstrates a little less efficacy (0.3471) but it is still better than other FPGA compatible methods like the YOLO+KF and Tiny-YOLO+KF algorithms. The discrepancy in results for the two datasets may be because of the nature of the captured scenes. Since our focus has been on developing and implementing adaptive subsampling-based tracking algorithms, we simply made use of the pre-trained networks forming the backbone of the object detectors. Training these networks on the training subset of the LaSOT dataset may have mitigated the discrepancies we are seeing from dataset to dataset. 

The ATOM and DiMP methods, although considered to be state-of-the-art in the field of tracking, do not perform well in the adaptive subsampling setup without the Kalman filter. Notice how the OTB100 score drops down to 0.2859 from 0.4625 for ATOM when we remove the Kalman filter. In a similar vein, the DiMP score for OTB100 drops down to 0.3398 from 0.4817 - quite a steep degradation. The same is true for the LaSOT dataset as well. This may be attributed to the fact that ATOM and DiMP heavily rely on target-specific information and scene details. When we enforce adaptive subsampling on the input data received by these trackers, there is a distinct dearth of information for the ATOM and DiMP networks to work with and they tend to make erroneous ROI predictions. Compare these to the algorithms where we use the same detectors but in combination with the Kalman filter. In the absence of fully sampled frames at every single time step, the Kalman filter provides tremendous assistance in keeping the trackers within the confines of the correct trajectory. Withdrawing this additional support from the filter makes it very difficult for the ATOM and DiMP methods to sustain their performance. Figure~\ref{fig:auc} demonstrates the success plots obtained for the various adaptive subsampling algorithms over a range of IoU thresholds. As expected, the greater the constraint on the IoU threshold, the worse the results obtained with the trackers. The tracking performance of these various methods can also be visualized in Figure~\ref{fig:comparison}, where the subsampling performance of the algorithms have been compared to the ground truth subsampling mask. The ECO+KF method does an excellent job of honing in on the object of interest while making sure to output a compact bounding box, implying large energy savings for the frame being shown in the figure.

\begin{table*}
\begin{center}
\begin{tabular}{l c c c c c c c c c c}
\hline
Dataset & MS+KF & YOLO\textsuperscript{~\cite{redmon2016you}}+KF & KCF\textsuperscript{~\cite{henriques2014high}}+KF & DAT\textsuperscript{~\cite{possegger15a}}+KF & ECO\textsuperscript{~\cite{danelljan2017eco}}+KF & ATOM\textsuperscript{~\cite{danelljan2019atom}} & DiMP\textsuperscript{~\cite{bhat2019learning}} & Tiny-YOLO+KF & ATOM+KF & DiMP+KF\\
\hline\hline
OTB100 &  0.2051 & 0.2709 & 0.2567 & 0.2573 & 0.4568 & 0.2859 & 0.3398 & 0.0809 & 0.4625 & 0.4817\\

LaSOT & 0.1928 & 0.2733 & 0.1809 & 0.1712 & 0.3471 & 0.2425 & 0.2942 & 0.1103 & 0.4282 & 0.4702\\

\hline
\end{tabular}
\end{center}
\caption{We report the AUC scores with IoU@[0:0.05:1] and keyframing interval of 11 on the two benchmarking datasets - OTB100 and LaSOT.}
\label{tab:main_results}
\end{table*}

\textbf{Keyframing:} As has been stated before in Section~\ref{sec:M}, we refer to the fully sampled images that the object detector receives as the key frames and the rest are referred to as subsampled frames. After the first fully sampled frame is fed through the object detector (update phase), we obtain the ROI location. For the consecutive subsampled frames, the Kalman filter utilizes the object detector output to update it's internal matrices and make ROI predictions until the object detector is activated again (prediction phase). As mentioned before, the keyframing interval is user-defined in this study. Note that a longer interval will have repercussions for tracking performance but will guarantee higher energy savings and faster computation. The interval can be adapted as per user needs.

\begin{figure*}[!t]
\begin{center}
   \includegraphics[width=0.85\linewidth]{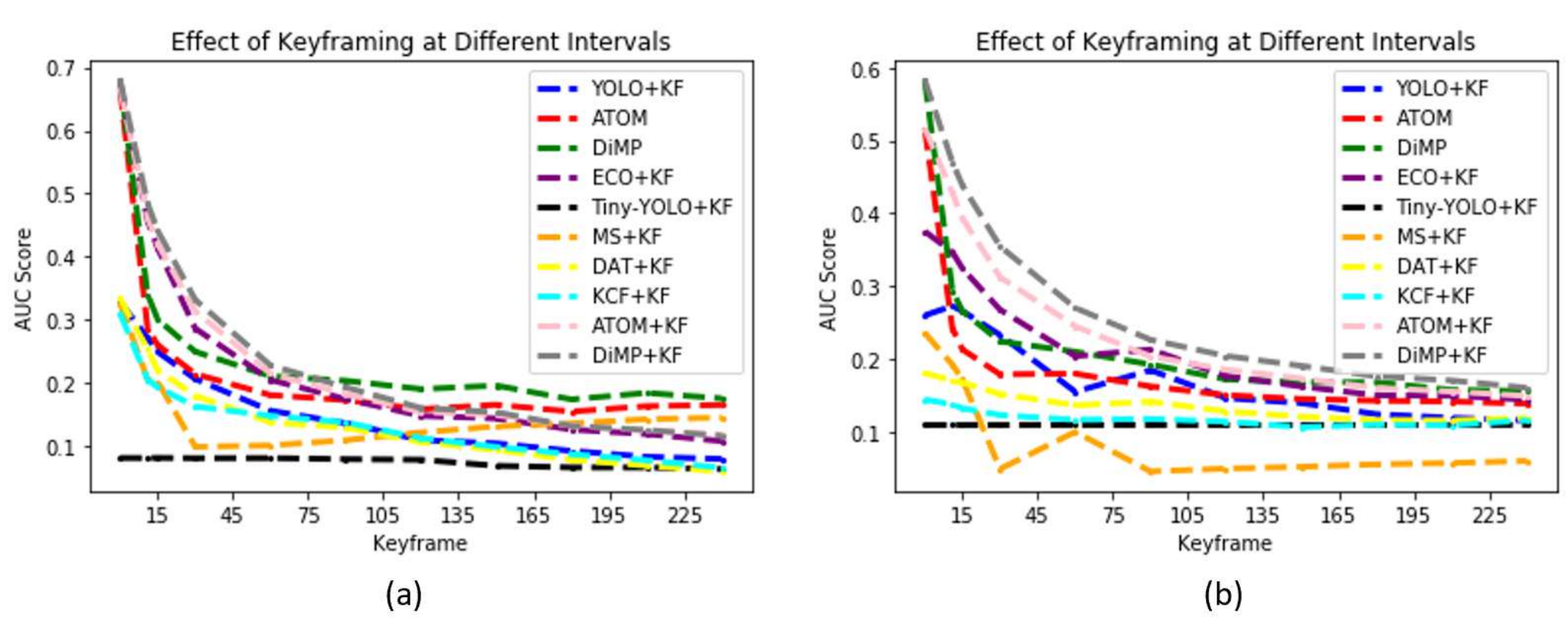}
\end{center}
\captionsetup{skip=-2mm}
   \caption{Results for the keyframing experiment. We have swept the keyframing interval from 0 to 240 for all of our adaptive subsampling algorithms on the (a) OTB100 dataset and the (b) LaSOT dataset and reported the auc score.}
\label{fig:keyframing}
\end{figure*}

Figure~\ref{fig:keyframing} clearly demonstrates the effect of increasing the keyframing interval. Comparing how the various algorithms are affected, it is evident that ECO+KF, ATOM+KF and DiMP+KF are the top candidates that are able to maintain significantly higher precision even at longer keyframing intervals. On the contrary, methods like the KCF+KF and especially the Tiny-YOLO+KF, which are shown to perform comparatively well at lower keyframing intervals, are not able to sustain that same degree of performance at higher intervals. In addition, note how the ATOM and DiMP methods minus the Kalman filter start deteriorating as the keyframing interval increases. Note that the methods don't all go to zero in AUC score as keyframing interval goes to infinity. We can attribute this anomalous behavior to the fact that the object detectors are given the first frame annotation to work with. Thus, the detectors have knowledge about the object appearance. Hence, even at higher keyframing intervals, the detectors are somewhat able to guess where the object may be located based on the object appearance information, albeit not as well as they would have done at shorter keyframing intervals as evident from Figure~\ref{fig:keyframing}. Providing completely random initial bounding box information would certainly throw these detectors, and then we would have seen the AUC scores go down to 0 with increasing keyframing intervals.  

Figure~\ref{fig:keyframing} depicts the effect of keyframing on the AUC score for intervals of up to 240 frames. For the most part, computer vision applications tend to aim for a latency of 30 FPS. It is promising that the ECO+KF, ATOM+KF and DiMP+KF methods show relatively stable performance at the 30 FPS mark.

\begin{table*}
\begin{center}
\begin{tabular}{l c c c c c c c c}
\hline
Dataset & MS & YOLO\textsuperscript{~\cite{redmon2016you}} & KCF\textsuperscript{~\cite{henriques2014high}} & DAT\textsuperscript{~\cite{possegger15a}} & ECO\textsuperscript{~\cite{danelljan2017eco}} & ATOM\textsuperscript{~\cite{danelljan2019atom}} & DiMP\textsuperscript{~\cite{bhat2019learning}} & Tiny-YOLO\\
\hline\hline
OTB100 &  0.2255 & 0.2725 & 0.1856 & 0.2217 & 0.3168 & 0.2141 & 0.2461 & 0.0804\\

LaSOT & 0.1991 & 0.2826 & 0.1637 & 0.1337 & 0.2961 & 0.2121 & 0.2493 & 0.1099\\

\hline
\end{tabular}
\end{center}
\caption{We replace the Kalman filter prediction and instead perform subsampling via memoization. We report the AUC scores with IoU@[0:0.05:1] and keyframing interval of 11 on the two benchmarking datasets - OTB100 and LaSOT.}
\label{tab:memo}
\end{table*}

\begin{figure*}[!t]
\begin{center}
   \includegraphics[width=0.85\linewidth]{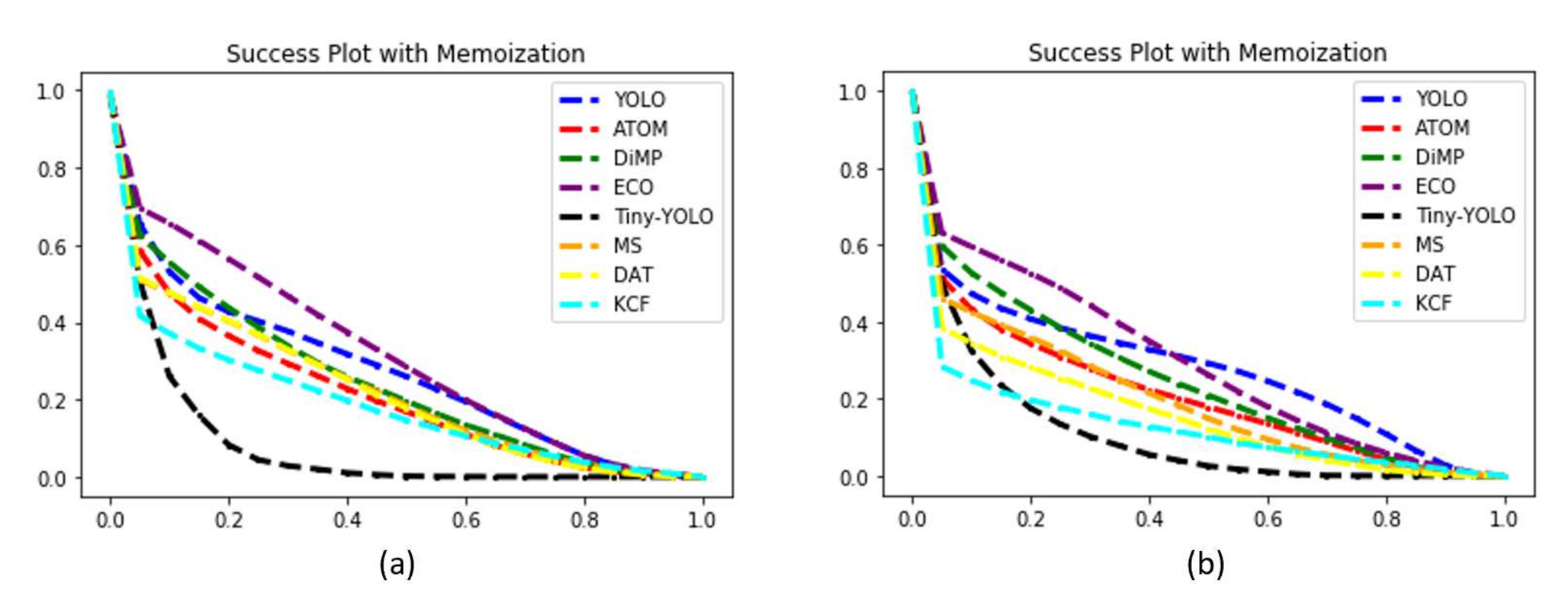}
\end{center}
\captionsetup{skip=-2mm}
   \caption{Results for the memoization experiment. We have swept the IoU thresholds from 0 to 1 (IoU@[0:0.05:1]) with a keyframing interval of 11 for all of our adaptive subsampling algorithms on the (a) OTB100 dataset and the (b) LaSOT dataset and reported the auc scores.}
\label{fig:memoization}
\end{figure*}

\textbf{Memoization:} 
Memoization is a known strategy in systems wherein the results of computationally expensive operations are cached for later use in the event that the same input signals are received in the future~\cite{acar2003selective, alvarez2005fuzzy, della2015performance, boos2016flashback}. 
We have shown results for ATOM and DiMP in the context of adaptive subsampling without Kalman filtering. This involves simply taking the output obtained for one image frame and using it to subsample the next frame, continuing the process until a keyframe is read out once again. Note the computational complexity of the process - an image needs to be processed by the network at every time step. To overcome this, we conduct another study where we treat the output of the keyframe as the ROI for all subsequent subsampled images. This setup mimics the systems memoization strategy, and we conduct this experiment for all of the object detectors that we have used for this study. We report the AUC scores we obtain with memoization in Table~\ref{tab:memo}. The ECO tracker outperforms the other methods on both the OTB100 and LaSOT datasets. However, the DiMP fares better when the Kalman filter is introduced to the pipeline in lieu of memoization. This shows that image statistics and scene content may have some bearing on tracker performance and that the ECO, ATOM and DiMP trackers are of similar calibre when it comes to tracking in this adaptive subsampling context. It is also interesting to note that withdrawing the support of the Kalman filter shows a decline in performance and this highlights the necessity of leveraging the Kalman filter to ensure greater tracking fidelity as well as lower latency and higher energy savings. Figure~\ref{fig:memoization} demonstrates the success plots for the memoization experiment on both the datasets. 

\textbf{Training on subsampled images:} Neural network-driven adaptive subsampling raises the question of how these trackers may perform if they are trained on subsampled images. The ATOM trained on subsampled images achieves an AUC score of 0.2001 on OTB100 and 0.1304 on LaSOT, where before (with the network trained on fully sampled images) it was achieveing an auc score of 0.4625 on OTB100 and 0.4282 on LaSOT. On OTB100, the DiMP performance deteriorates down to an auc score of 0.1065 from 0.4817 and on LaSOT it degrades down to 0.0758 from 0.4702. This can be attributed to the fact that reducing scene information at training time reduces the learning capabilities of these networks and they fail to predict correct object trajectories at test time.





\begin{figure*}[!t]
\begin{center}
   \includegraphics[width=0.9\linewidth]{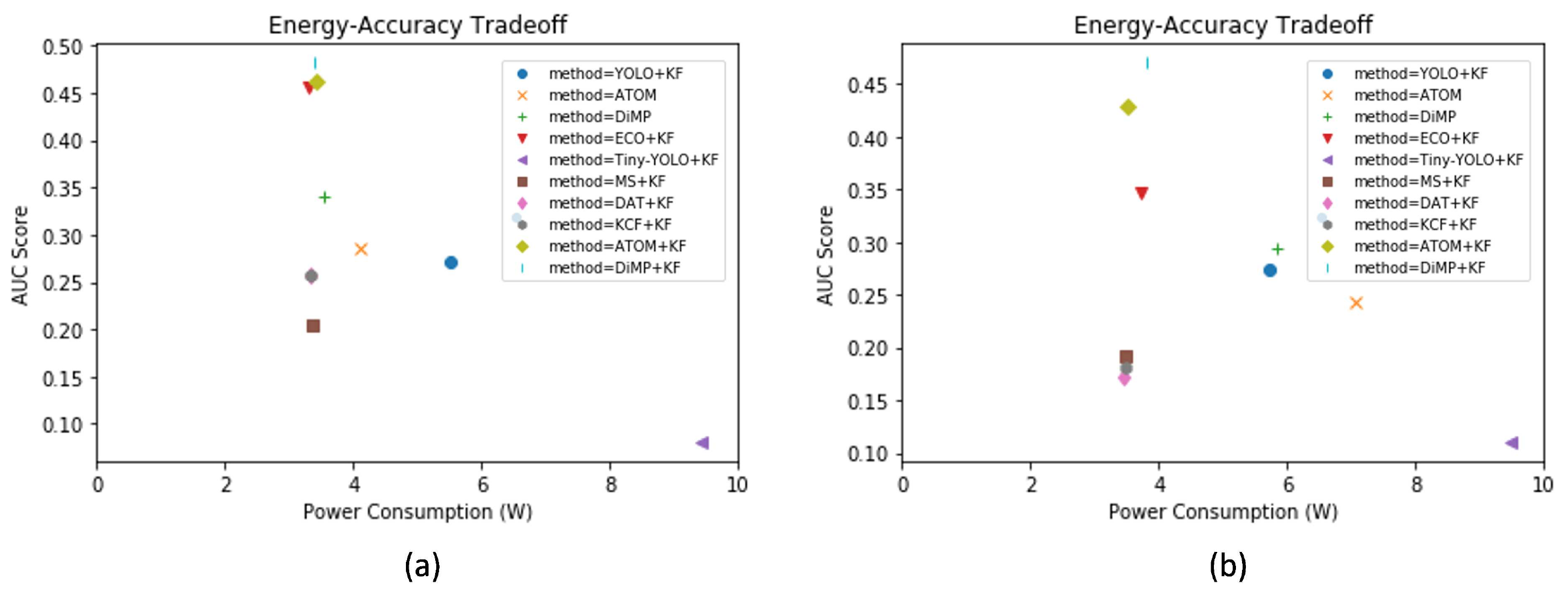}
\end{center}
\captionsetup{skip=-2mm}
   \caption{Scatter plot depicting the AUC score vs. power consumption tradeoff for the (a) OTB100 and (b) LaSOT dataset (with a keyframing interval of 11). Aside from being FPGA-compatible, the ECO tracker demonstrates high-precision tracking performance while being comparable to the other methods in terms of power consumption.}
\label{fig:tradeoff}
\end{figure*}
\textbf{Power Analysis:} To analyze the power savings that can be achieved with our selected approaches, we conduct a study where we characterize the energy requirements of several CMOS image sensors based on the analysis done in~\cite{likamwa2013energy}. The algorithm generated bounding boxes are utilized to prompt the image sensor to skip columns not associated with the ROI during frame read out. The idea is to mitigate power consumption by reading out fewer pixels. Sensors B$1$, B$2$ and B$3$ from~\cite{likamwa2013energy} come with resolution $3264$x$2448$, $2592$x$1944$ and $752$x$480$ respectively. 
The power analysis model equations reveal that the average power consumption is proportional to the image resolution~\cite{likamwa2013energy}: 
\begin{equation}
\label{eq:power}
    P = \frac{P_{idle}T_{exp} + P_{active}T_{active}}{T_{frame}}\\
\end{equation}
\begin{equation}
\label{eq:power2}
    P = \alpha_{1}.R.T_{exp}.f + \frac{R.c_{2}.N}{f}
\end{equation}
where $R$ represents frame rate (fixed at $30$ fps), $T_{exp}$ is the exposure time (fixed at $0.05$ms), $N$ represents frame resolution, $c_{2}$ denotes static power consumption (fixed for every sensor: B$1$: $159.0$, B$2$: $93.0$ and B$3$: $13.1$), $\alpha_{1}$ (fixed for every sensor: B$1$: $4.0E-06$, B$2$: $8.2E-07$ and B$3$: $3.35E-06$) is a sensor intrinsic independent of resolution and $f$ represents the optimal clock frequency dependent on resolution $(\frac{c_{2}.N}{\alpha_{1}.T_{exp}})^{1/2}$.


Figure~\ref{fig:tradeoff} visualizes the power-accuracy tradeoff of the various methods, where accuracy is denoted by the AUC score on the OTB100 and LaSOT datasets and corresponding power consumption is given in watt. As is evident from the tradeoff plot, the ECO-based approach manages to attain comparable tracking performance to the best candidates - ATOM+KF and DiMP+KF - while retaining similar power consumption levels as the more energy-efficient trackers. Given its FPGA compatibility and its performance with respect to the other neural network-based approaches, we selected the ECO tracker for hardware acceleration. Case in point, the YOLO and tiny-YOLO object detectors - although very easy to port to the FPGA - are less efficient in terms of both power consumption and tracking precision in comparison to ECO.

\section{Hardware Acceleration for FPGAS.}
\label{sec:hardware}

The previous section focused on the analysis and performance evaluation of our adaptive subsampling algorithm in software. The purpose of this section is to discuss the hardware implementation of these algorithms, specifically targeting an embedded device such as a field programmable gate array (FPGA).   \newline

Deep learning tasks are usually performed on general purpose computation devices such as microcontrollers and graphic processing units (GPUs) which provide a high degree of flexibility but have high energy consumption. 
On the other hand, application specific integrated circuits (ASICs) are devices dedicated to a single specific purpose that provide ultimate area and energy efficiency at the cost of losing all flexibility~\cite{DL_FPGAS}.
A FPGA is a set of 2D reconfigurable resources that allow mapping of custom hardware architectures. Given their reconfigurable nature, they provide the flexibility of general purpose devices while granting the ability to apply spatial and temporal parallelism, retaining the area and energy efficiency of ASICs~\cite{FPGA_tradeoff}.
FPGAs are also a popular alternative to move IoT applications from centralized cloud-computing environments towards geographically located edge-computing servers~\cite{edge_FPGA}.\newline

In the past, the implementation of algorithms on FPGAs required the knowledge of Hardware Description Languages (HDL) as Verilog or VHDL.  This prerequisite prevented software and algorithm developers from taking advantage of this technology.
However, in recent years several open-source (OpenCL) and proprietary tools (Xilinx Vivado HLS and Vitis, Intel HLS Compiler) have been developed  that allow high-level synthesis of applications from high-level programming languages like C++ and Python. In a recent survey~\cite{FPGA_survey}, an exhaustive list of current and abandoned FPGA HLS tools is presented which is useful reference for the reader. \newline

\subsection{Hardware Implementation}
To accelerate the deep learning component of the object trackers in our adaptive subsampling algorithms we used Xilinx’s HLS environment Vitis AI~\cite{vitis_ai}. This is a stack of development tools that allow AI inference on Xilinx FPGAs. It supports mainstream frameworks such as PyTorch and TensorFlow which allow the development of deep learning applications.
Xilinx Ultrascale+ MPSoC edge devices include a processing system (ARM core) and programmable logic (PL) on the same chip.  
Xilinx provides a synthesized deep learning processing unit (DPU) that maps onto the FPGA, along with a PetaLinux environment that allows the execution of Python scripts and open-source libraries directly on the embedded system.  Vitis AI provides Python APIs that allow communication between the programmable logic and processing system. 
To further accelerate other non-deep-learning components of our algorithm, other tools like the Vitis software platform or Vitis libraries are needed.  However, these tools require writing and synthesis of C++ kernels along with the DPU, adding complexity to the system.  Given the complexity of this approach, we chose to only accelerate the deep learning section of our algorithm using Vitis AI, while running the rest of our code using standard Python libraries.
\newline
\vspace{-0.3cm}
\begin{figure}[h]
    \centering
    \includegraphics[width=.5\textwidth]{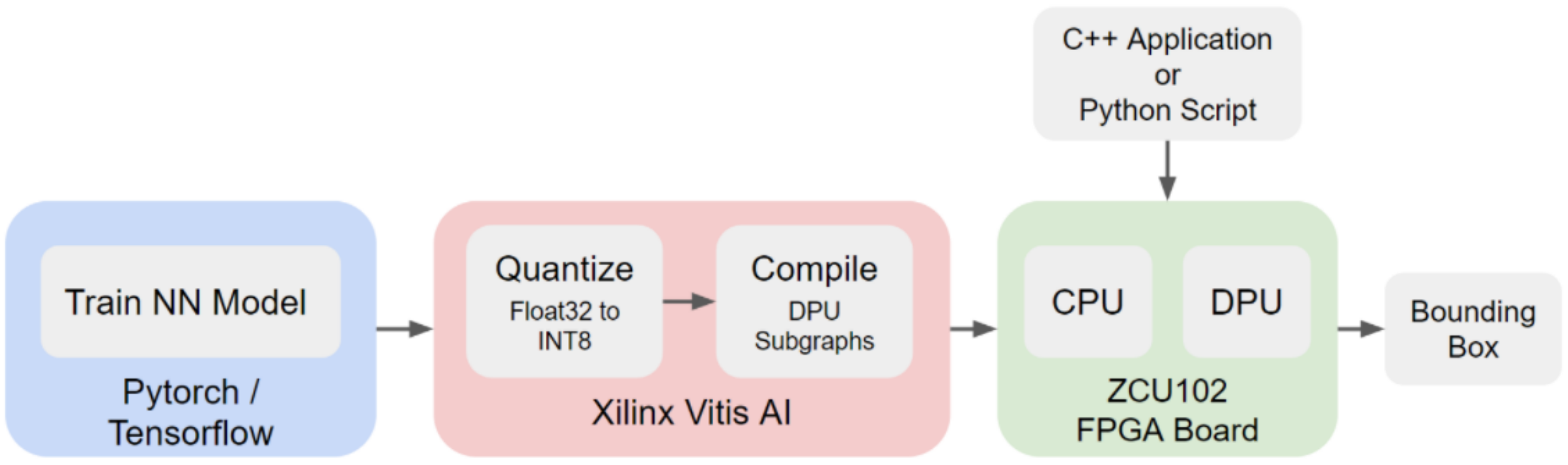}
    \caption{Vitis AI flow.}
    \label{fig:vaiflow}
\end{figure}
Figure~\ref{fig:vaiflow} shows how a deep learning application is deployed on an Edge FPGA using Vitis AI.  First, the model is defined and trained offline using one of the supported deep learning frameworks.  
Next, VAI Quantizer converts weights and activations from 32-bit floating-point to 8-bit fixed point.  This reduces computation complexity and memory bandwidth of the model.
The quantized model is then passed into the VAI Compiler which maps the network model into a graph-based optimized instruction sequence based on the DPU architecture. The output is a compiled \textit{.xmodel} which will be invoked to run in the programmable logic by VAI Python APIs~\cite{vai_ug}.
One disadvantage of this method is that the input and output size of the compiled models are fixed, therefore we must preprocess our input before sending it for processing to the DPU. \newline

\begin{figure}[h]  
    \centering
    \includegraphics[width=.3\textwidth]{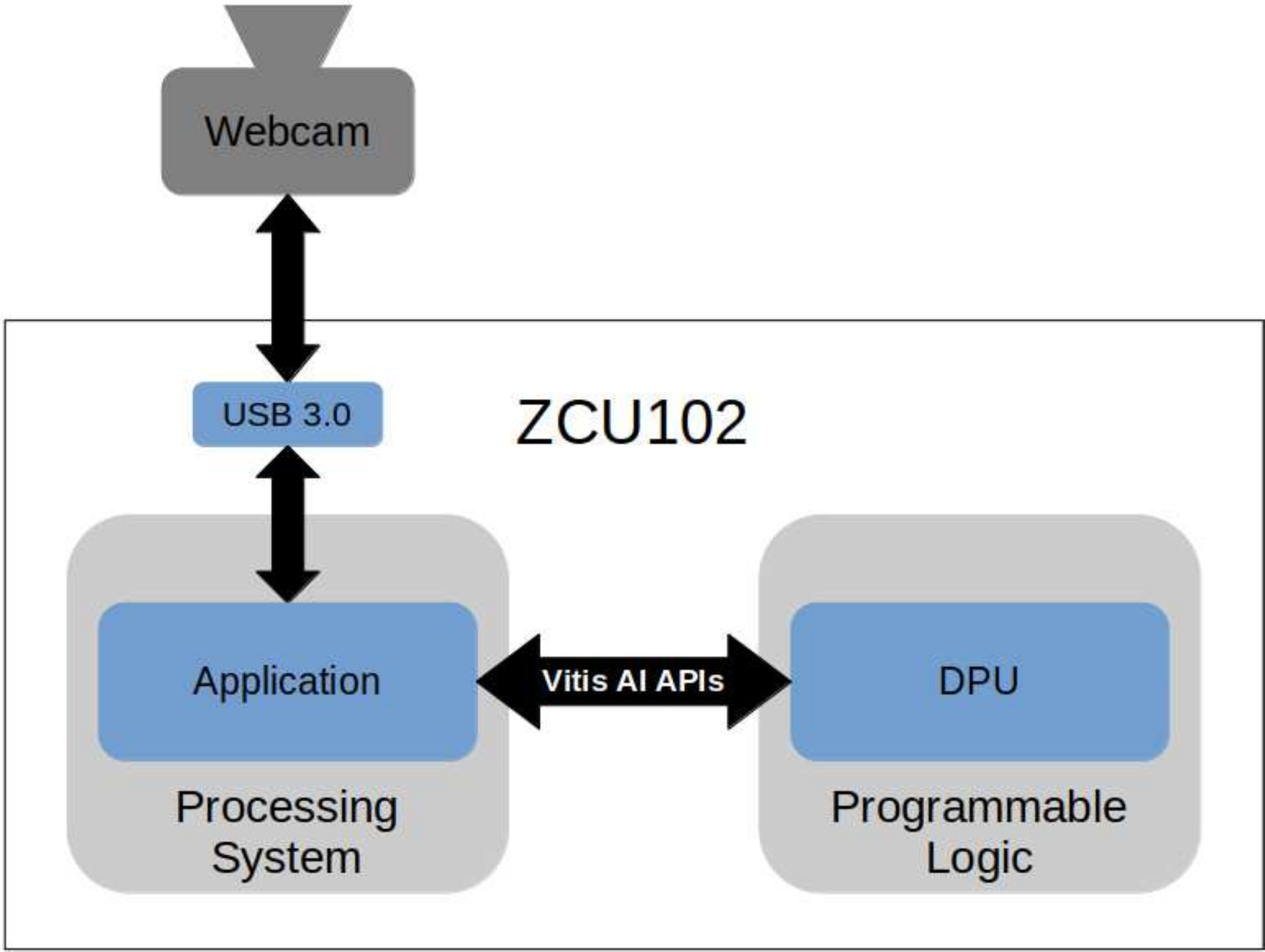}
    \caption{ System diagram.  The webcam is connected via USB to the board.  Our application runs on the processing system while the DPU is mapped on the programmable logic.  Vitis AI APIs allow communication between both of them.}
    \label{fig:hwdiagram}
\end{figure}

\begin{figure}[h]
    \centering
    \includegraphics[width=.3\textwidth]{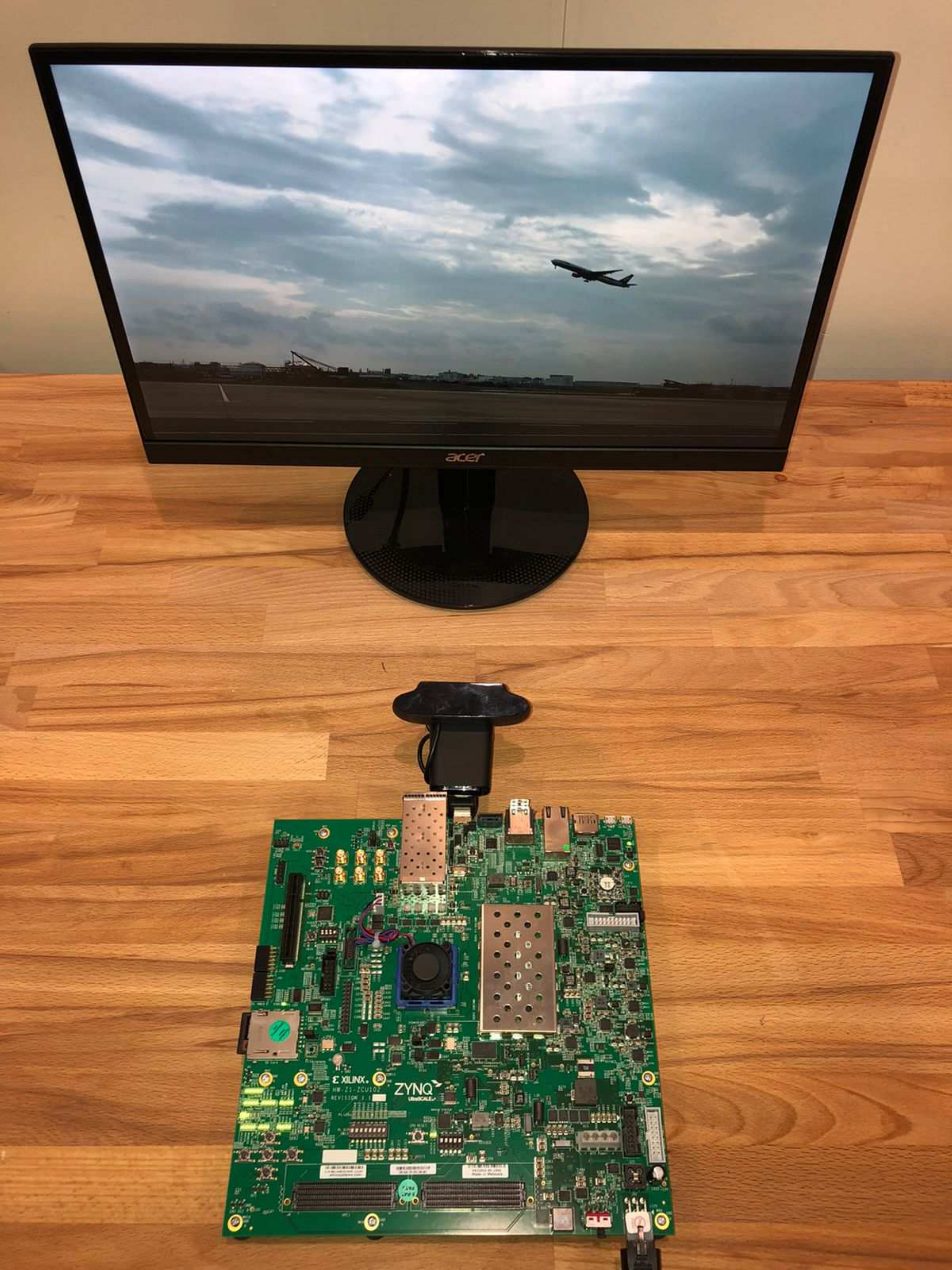}
    \caption{ Our hardware setup: A sequence is displayed on an Acer LCD monitor.  A Logitech C920 webcam connected to a ZCU102 board captures an image on every keyframe to perform object detection.  Our system emulates ROI capture by predicting the bounding box every subsequent frame.  This information would be sent to an ROI camera to perform adaptive subsampling.}
    \label{fig:hwsetup}
\end{figure}

Our hardware setup consists of a Xilinx ZCU102 evaluation board,  and a Logitech C920 HD camera connected by USB.  The PetaLinux environment is loaded via an SD card, which contains the hardware bitstream, software libraries and application scripts to run the experiments. Figure~\ref{fig:hwdiagram} and Figure~\ref{fig:hwsetup} display a system diagram and our hardware setup respectively.
Within our application, we used OpenCV to capture frames from the camera.  Our script also preprocessed the captured frames to fit the model’s required input size.  
We also explored the use of ROI-capable cameras to complete our system.  However, we were not able to find a camera that was able to dynamically adjust its ROI while being compatible with our FPGA evaluation board.  Given this, we chose to digitally simulate ROI for demonstration purposes as performed in other papers in the literature~\cite{rhythmic_pixels}. The image capture, preprocessing, digital ROI, Kalman filter update and prediction steps, and any other postprocessing required by each specific tracker is done in the processing system of the FPGA board. \newline
\vspace{-0.3cm}

While ATOM and DIMP trackers performed slightly better than ECO, they made use of custom layers to perform feature extraction.  These layers are not compatible with Vitis AI and would require major efforts and HDL knowledge to accelerate and deploy on our system.  
ECO tracker uses VGG \cite{vgg} (a classical convolutional neural network architecture) to perform feature extraction.  Given its similar performance and layer compatibility we chose to accelerate the ECO + KF algorithm.
To implement this algorithm on our system, we accelerated the feature extraction module of ECO following the flow previously described and displayed in Figure \ref{fig:vaiflow}.  This allowed us to run the deep learning component of our algorithm in the programmable logic.  The PetaLinux environment of our system allowed us to install and take advantage of popular Python libraries like PyTorch and OpenCV to perform several pre and post-processing steps on the processing system.  It is worth mentioning that some of the libraries used by the original implementation of ECO were not compatible with the ARM core of our board.  This forced us to find alternative libraries and perform substitutions to be able to run this tracker.  
We also chose to accelerate YOLOv3 + KF given that VAI model zoo \cite{modelZoo} already provides a pre-trained YOLOv3 model (\textit{yolov3\_tf\_voc}) and postprocessing code \cite{yolov3DNNDK}.  We incorporated this code into our application and added required camera capture, preprocessing and postprocessing steps, as well as the Kalman filter implementation and digital ROI simulation.  
In our previous work we also accelerated the MS+KF algorithm using SDSoC and xfOpenCV accelerated libraries~\cite{9191146}.

\subsection{Hardware Results}
In Table \ref{tab:hw_results} we show the performance of our algorithm and system.  Our algorithm performance takes into consideration the delay of capturing an image, performing preprocessing, detection, postprocessing and updating the Kalman filter on every keyframe.  While only executing the kalman prediction step on subsequent frames.  
YOLO + KF achieves a performance of 65 frames per second, while ECO reaches a speed of 19 frames per second.  Although ECO + KF is slower, from the experiments in Table~\ref{tab:main_results} we know that it has a better tracking performance than YOLO + KF.  We can therefore conclude that there is a tradeoff between speed and tracking performance of these two algorithms.  

Our system performance adds the image capture delay on each subsequent frame.  In our current system, this capture delay is identical to the capture delay during keyframes.  In a real ROI system, the capture delay of the subsequent frames would be reduced given that we would only need to capture a reduced set of pixels. 
In this metric, YOLO + KF achieves a performance of 24 frames per second while ECO  reaches a speed of 13 frames per second. 

\begin{figure}[h]
    \centering
    \includegraphics[width=.45\textwidth]{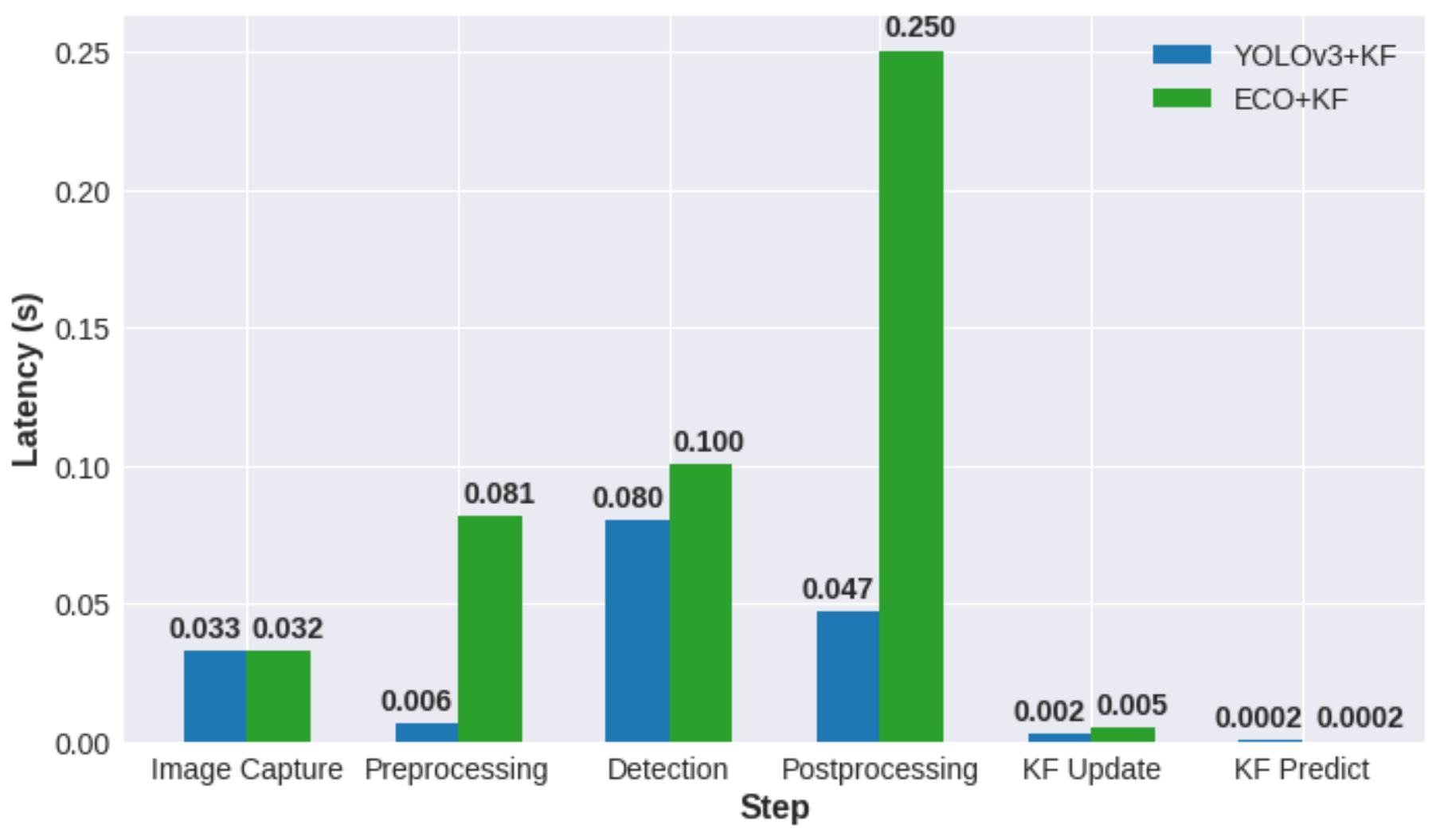}
    \caption{Breakdown of algorithm performance on hardware.  Image capture, preprocesing, detection and postprocessing only occur every keyframe.  On subsequent frames the Kalman Filter predicts the next bounding box.}
    \label{fig:latency_breakdown}
\end{figure}

In Figure~\ref{fig:latency_breakdown} we show a breakdown of the latency at each step for both algorithms.  We can see how the image capture, dpu detection and kalman filter steps have a similar latency in both algorithms.  However, the preprocessing and postprocessing latency of the ECO tracker is significantly higher than the one from ECO.  This is due to the transforms and optimizations that ECO requires to do its processing.  These operations are not trivial to implement on an FPGA using the know HLS tools.  But since these steps are only done every keyframe, their effect is diminished by using the Kalman Filter to predict the bounding box on subsequent frames.  


\begin{table}[h]
    \begin{center}
    \begin{tabular}{l c c c c c c c c c c}
    \hline
    Algorithm & Algorithm FPS & System FPS & Keyframe \\
    \hline\hline
    ECO\textsuperscript{~\cite{danelljan2017eco}}+KF  & 19.23 & 13.42 & 10\\
    
    YOLO\textsuperscript{~\cite{redmon2016you}}+KF & 65.39 & 24.6 & 10\\
    
    \hline
    \end{tabular}
    \end{center}
    \caption{Performance of our algorithm in hardware.}
    \label{tab:hw_results}
\end{table}

The performance of the object detectors can be further increased by multithreading techniques.  However, this is only achieved by pre-capturing images so the DPU can work on multiple frames at the same time.  For comparison, the YOLOv3 model provided by Xilinx can achieve 34.5 FPS (albeit without camera capture, pre and post-processing)\cite{modelZoo}.  To increase our throughput we would need to multithread the camera capture pre and post-processing, which is not compatible with our algorithm.

\begin{figure}[h]  
    \centering
    \includegraphics[width=.15\textwidth]{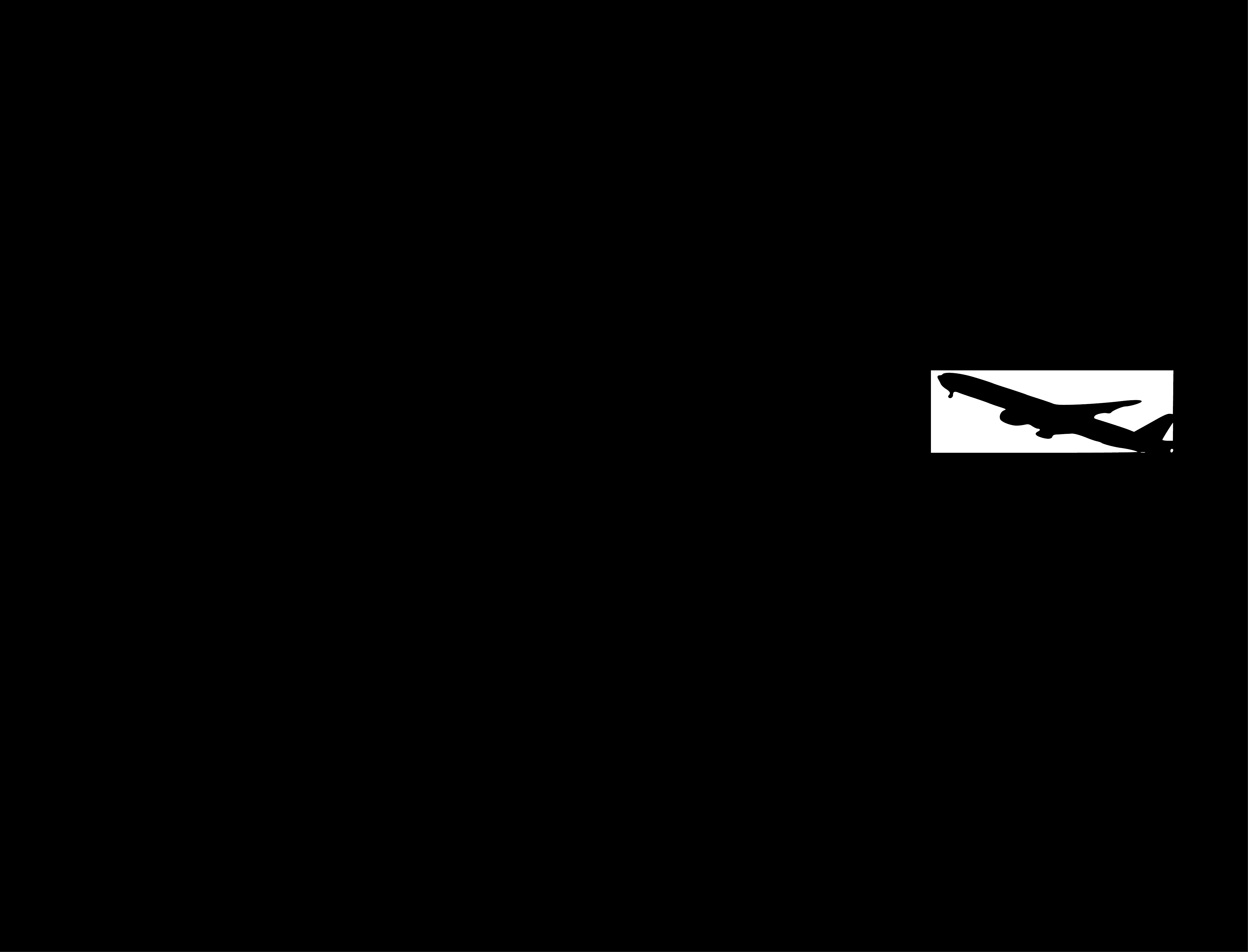}
    \includegraphics[width=.15\textwidth]{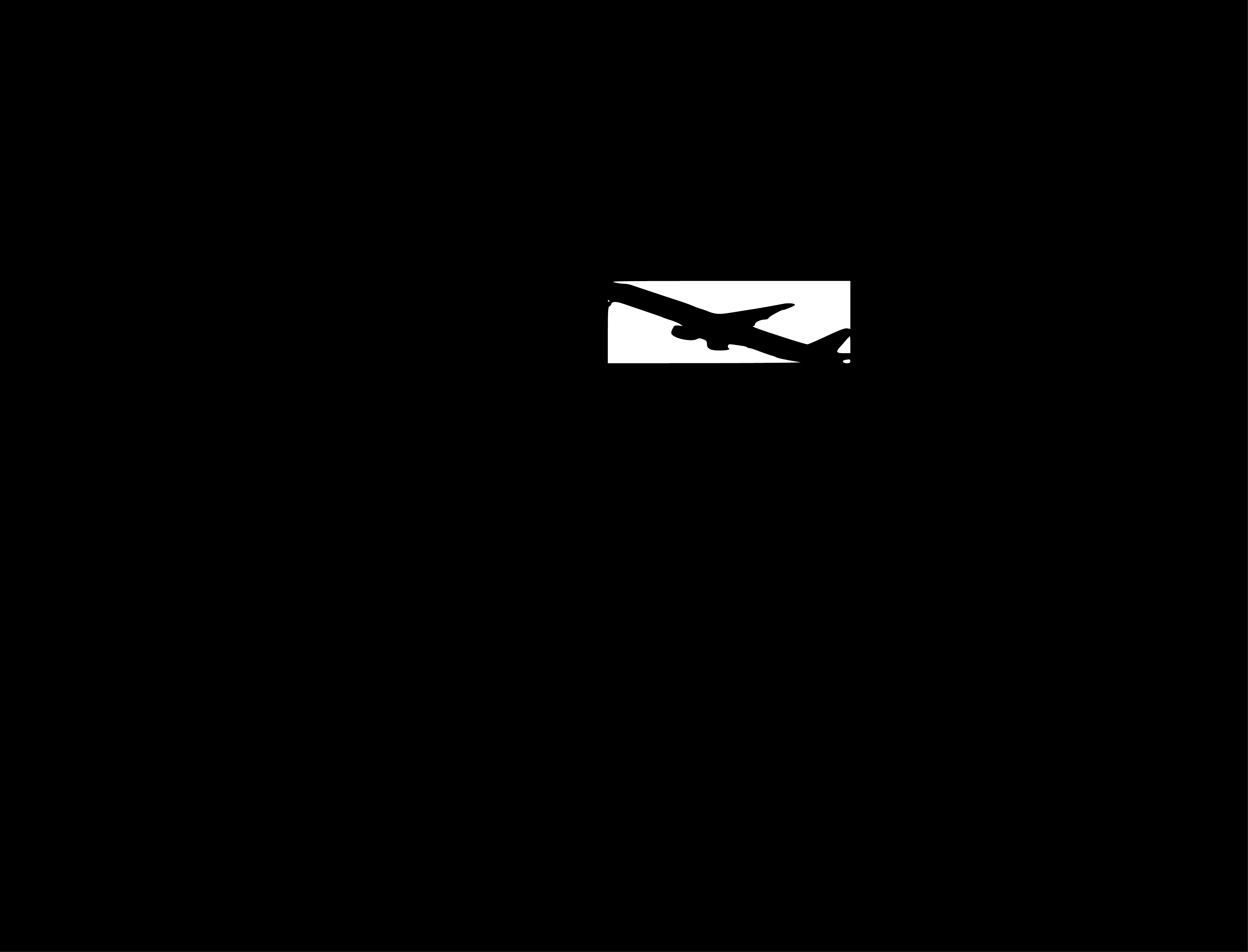}
    \includegraphics[width=.15\textwidth]{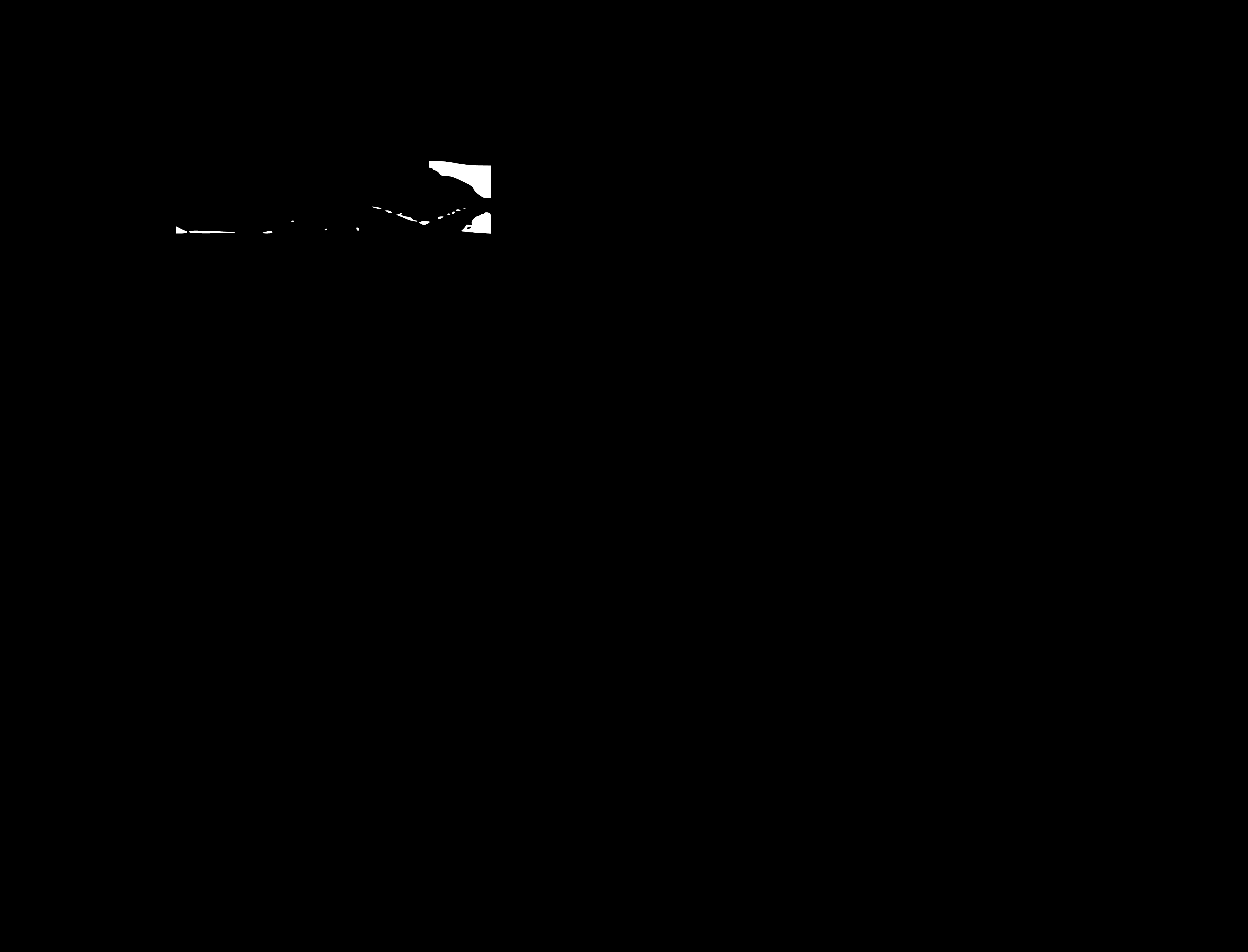}\\
    \vspace{.1 cm}
    \includegraphics[width=.15\textwidth]{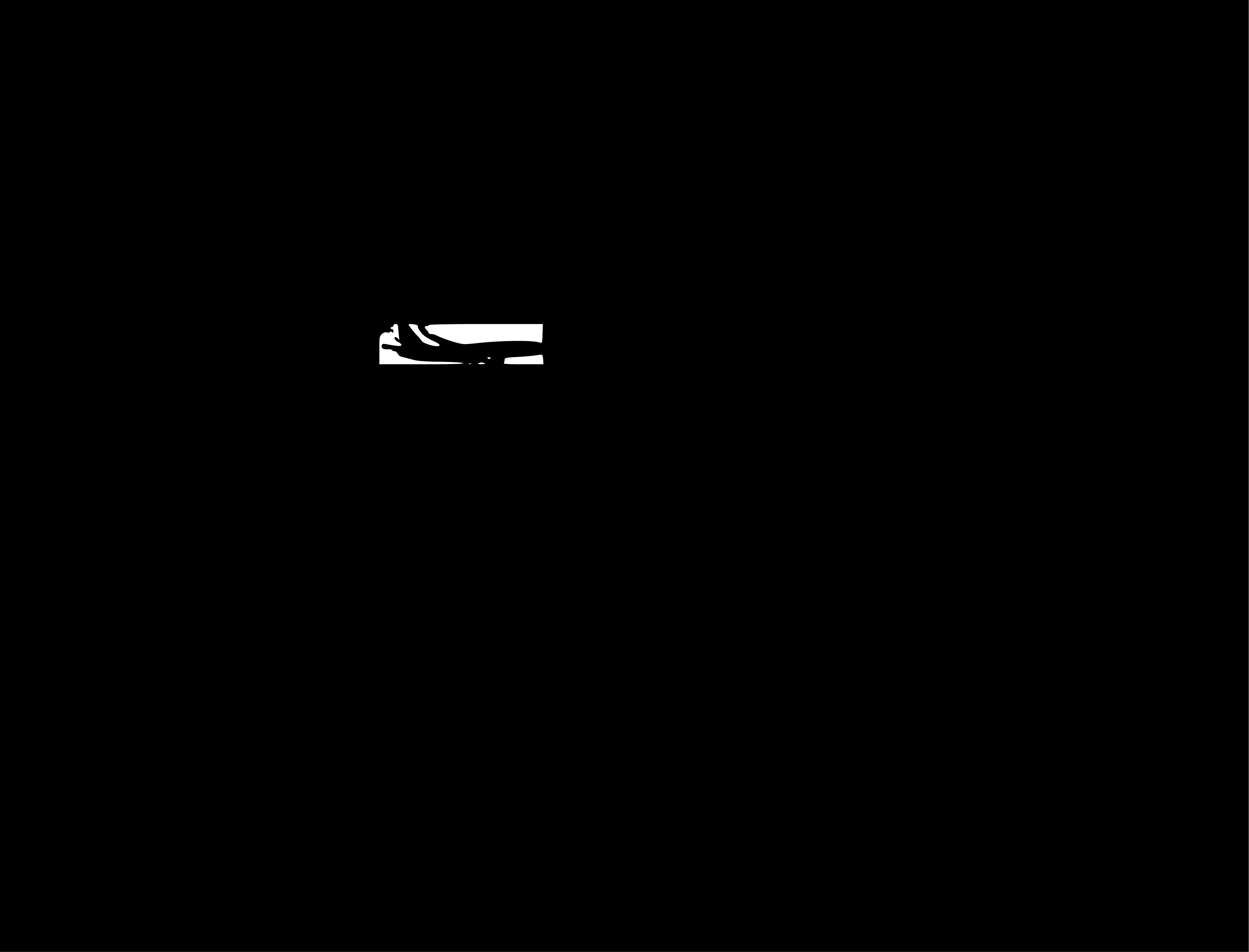}
    \includegraphics[width=.15\textwidth]{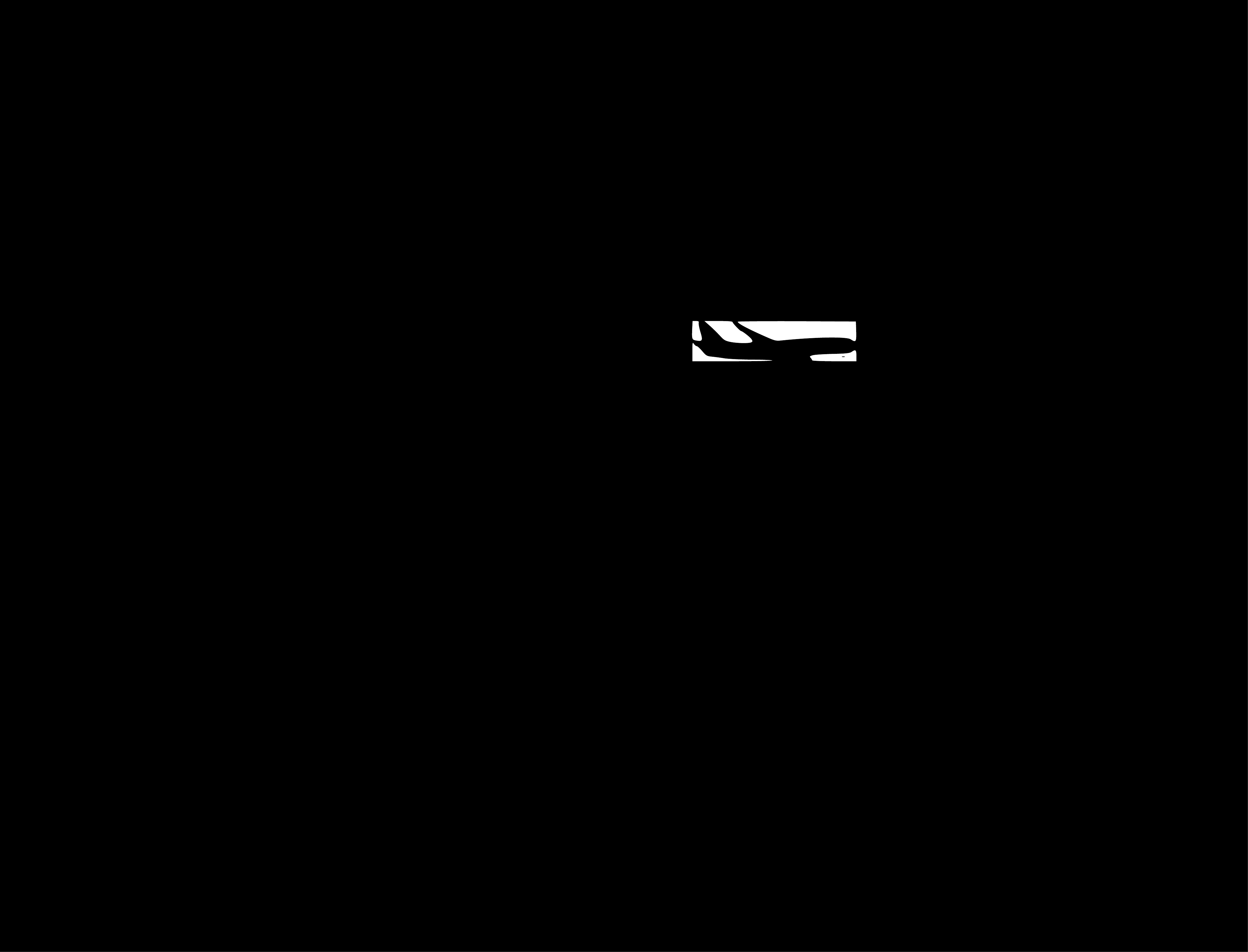}
    \includegraphics[width=.15\textwidth]{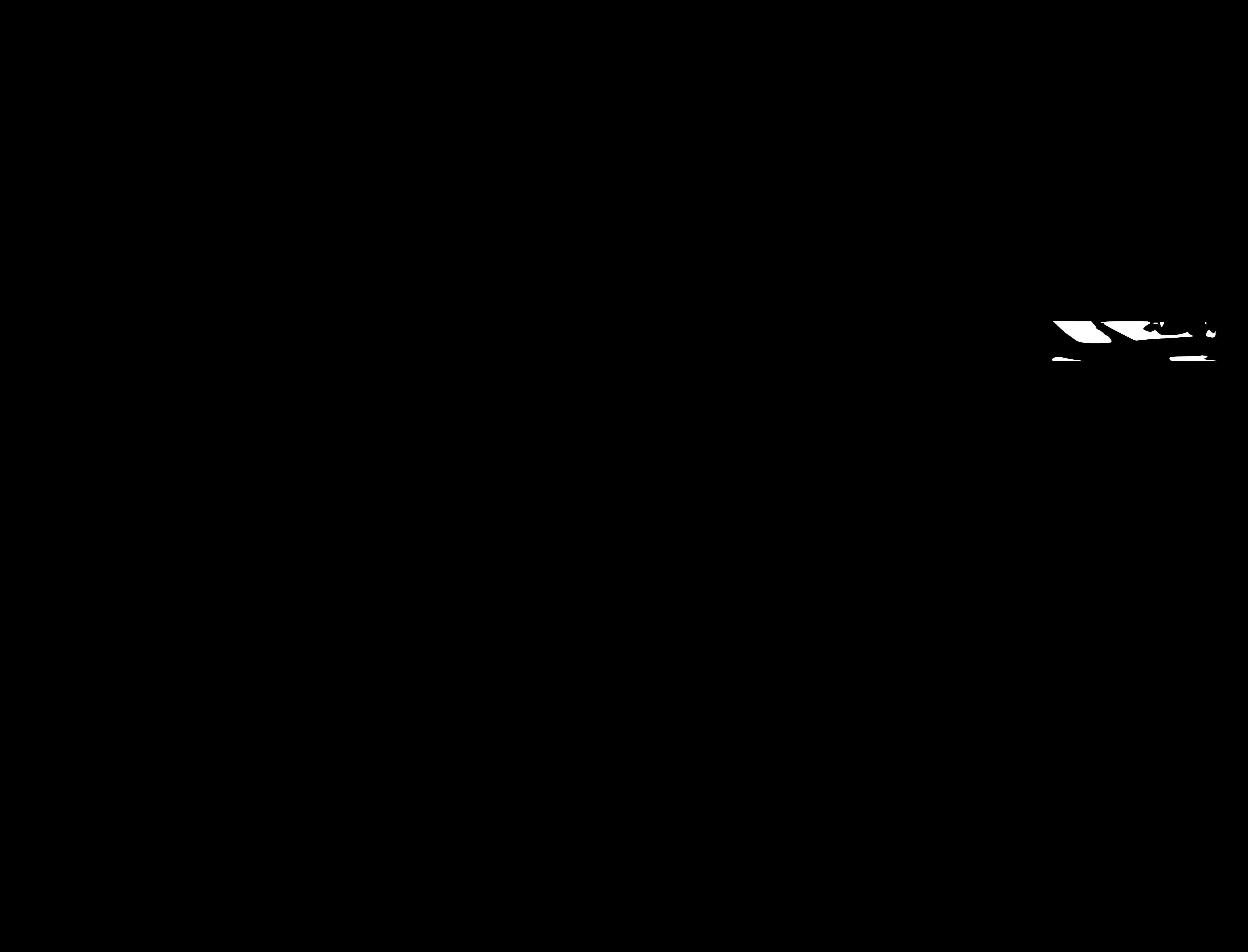}
    \caption{ROI simulation: We simulate adaptive subsampling by masking the area outside the computed ROI.  The ROI is obtained by performing object detection every keyframe and Kalman filter prediction every subsequent frame.  
    } 
    \label{fig:roiex}
\end{figure}

\begin{figure}[h]
    \centering
    \includegraphics[width=.45\textwidth]{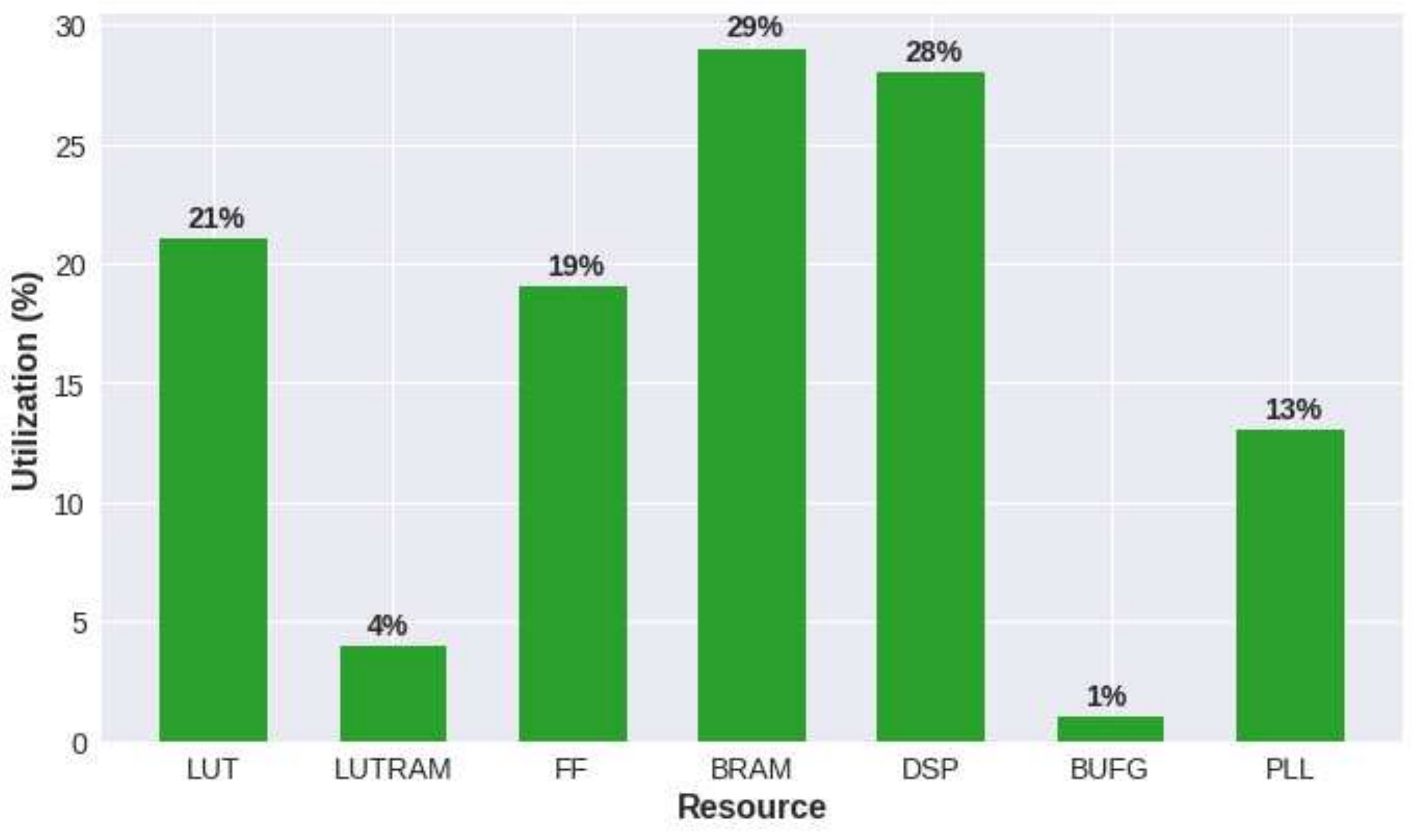}
    \caption{Resource Utilization (\%) of a single core DPU architecture.}
    \label{fig:resutil}
\end{figure}

As previously mentioned, the DPU is a highly optimized module that performs neural network computations.  It is designed to be resource-efficient while performing neural network inference. The DPU has several configuration parameters that can be customized for any specific application.  For both of our algorithms, we elected to use the same single-core, 4096-architecture DPU.  This configuration uses the fewest possible resources without sacrificing performance on a single-thread.  As shown in Figure~\ref{fig:resutil} and Table~\ref{tab:resourceUtil}, the DPU uses fewer than 30\% of every category of resources.  This allows room for other accelerated kernels to be mapped simultaneously on the programmable logic of our system. 

\begin{table}[h]
    \centering
    \begin{tabular}{lrr}
    \hline
    Resource & \multicolumn{1}{l}{Utilization} & \multicolumn{1}{l}{Available} \\ 
    \hline \hline
    LUT    & 58,734  & 274,080 \\
    LUTRAM & 6,226   & 144,000 \\
    FF     & 106,294 & 548,160 \\
    BRAM   & 261    & 912    \\
    DSP    & 704    & 2,520   \\
    BUFG   & 3      & 404    \\
    PLL    & 1      & 8      \\ \hline
    \end{tabular}
    \caption{Used and available resources on the ZCU9EG FPGA from our ZCU102 evalution board.}
    \label{tab:resourceUtil}
\end{table}




\section{Discussion}

This work paves the way for future research in adaptive subsampling-based tracking which has tremendous potential for energy optimization of image sensors. Coupling off-the-shelf object detectors with a Kalman filter results in an efficient mechanism for reconfiguring image sensors on the fly by reasoning about future object trajectories. We identify an ideal candidate out of these adaptive subsampling algorithms and map it onto an FPGA to demonstrate the feasibility of implementing such methods with a goal to maximizing the energy efficiency. 

There is a lot of scope for expanding this work given the current limitations of the proposed approach. Firstly, the predictor in our current framework is a simple Kalman filter. Switching this out with a powerful, state-of-the-art predictor may help improve the tracking precision by leaps and bounds. In addition, the Kalman filter being a classical state estimator requires frequent correctional measurements from an external source (the object detectors in our case). Alternatively, the power of neural networks can be leveraged here to attain superior tracking performance with a network-based predictor. Our intuition is that we may be able to operate at high keyframing intervals with a neural network without compromising the tracking fidelity. Another limitation of our current work has to do with the novelty of the technology. Xilinx has only recently come out with their Vitis AI toolflow and it is still very much in its early stages. We were bottlenecked by the limitations of the Vitis software in implementing the ATOM and DiMP-based approaches. Programming in the arm core and DPU and transferring data back and forth between the two processing units was also challenging for our particular use case. It required piece-by -piece investigation and restructuring of the modules in the ECO tracker. 

Finally, adaptive subsampling and ROI technology are exciting new areas of research with tremendous potential. In this work, we have shown FPGA acceleration as a mode of implementing adaptive subsampling. There is potential to explore ASICs in this area as well in lieu of FPGAs in order to attain better latency and optimized performance. The final step in this avenue of research would be integrate such custom adaptive subsampling algorithms with a real ROI-capable camera sensor, and study the real life performance in terms of latency and precision.

\section*{Acknowledgment}
This work was supported by NSF CCF-1909663, the SenSIP Center at ASU, Air Force Phase II STTR F19A-015-0183 and a gift from Qualcomm, Inc. This work has been approved for public release (AFRL-2021-4330); distribution is unlimited. We also thank Esen Salcin, Kostas Moutafis, Lloyd Linder, Sai Medapuram, Doug Bane and Matt Engelman from Alphacore for useful discussions. 

\ifCLASSOPTIONcaptionsoff
  \newpage
\fi

\bibliographystyle{IEEEtran}
\bibliography{IEEEexample.bib}



\vspace{-1.5cm}
\begin{IEEEbiography}[{\includegraphics[width=1in,height=1.25in,clip,keepaspectratio]{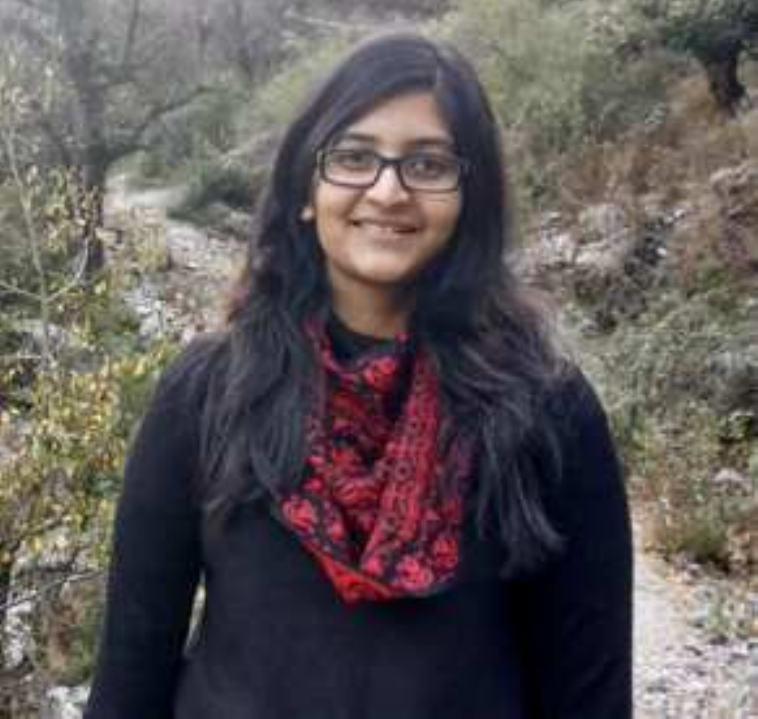}}]{Odrika Iqbal}
Odrika Iqbal is a 2nd year Ph.D. student at Arizona State University in the School of Electrical, Computer and Energy Engineering (ECEE). She is co-advised by Dr. Jayasuriya and Dr. Andreas Spanias. Odrika received her BSc degree in EEE at Bangladesh University of Engineering and Technology in September 2017. Her research interests are in computational imaging, machine learning and computer vision.
\end{IEEEbiography}

\vspace{-1.5cm}
\begin{IEEEbiography}[{\includegraphics[width=1in,height=1.25in,clip,keepaspectratio]{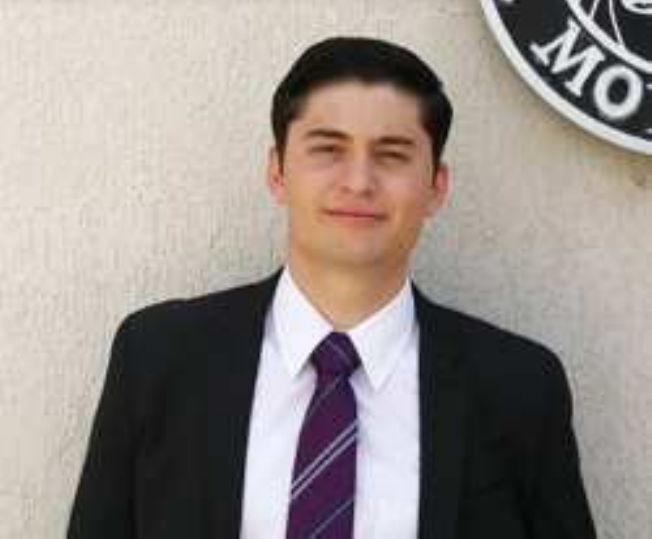}}]{Victor Torres}
Victor Torres is a 1st year MS student at Arizona State University in Computer Engineering. He received his BEng from Instituto Tecnológico y de Estudios Superiores de Monterrey (ITESM) in 2020. Victor has received the Fulbright-García Robles Scholarship from the Mexico-United States Commission for Educational and Cultural Exchange. His research interests are in embedded systems and artificial intelligence.
\end{IEEEbiography}
\vspace{-1.5cm}
\begin{IEEEbiography}[{\includegraphics[width=1in,height=1.25in,clip,keepaspectratio]{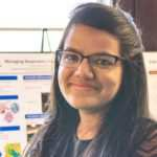}}]{Sameeksha Katoch}
Pursuing Ph.D. in Electrical Engineering from ASU specializing in Computer Vision focusing on aspects of  Time Series Prediction. My research interests include Image Processing, computer vision, and machine learning. Currently working on CPS project (Synergy: Image Modelling and Machine Learning Algorithms for Utility- Scale Solar Panel Monitoring) with the Sensor Signal and Information Processing (SenSIP) Lab. Working on dynamic texture synthesis utilizing both classical machine learning algorithms and state-of-art deep learning paradigms.  Consequently, these results will be used in Solar PV power forecasting which is useful to integrate the solar power in the existing power grids.
\end{IEEEbiography}

\vspace{-1.5cm}
\begin{IEEEbiography}[{\includegraphics[width=1in,height=1.25in,clip,keepaspectratio]{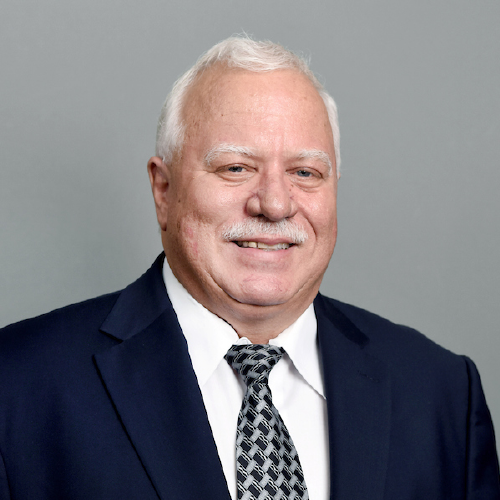}}]{Andreas Spanias}
Andreas Spanias is Professor in the School of Electrical, Computer, and Energy Engineering at Arizona State University (ASU). He is also the director of the Sensor Signal and Information Processing  (SenSIP) center and the founder of the SenSIP industry consortium (also an NSF I/UCRC site). His research interests are in the areas of adaptive signal processing, speech processing, machine learning and sensor systems. He and his student team developed the computer simulation software Java-DSP and its award-winning iPhone/iPad and Android versions. He is author of two textbooks: Audio Processing and Coding by Wiley and DSP; An Interactive Approach (2nd Ed.). He served as Associate Editor of the IEEE Transactions on Signal Processing and as General Co-chair of IEEE ICASSP-99. He also served as the IEEE Signal Processing Vice-President for Conferences. 
\end{IEEEbiography}

\vspace{-1.5cm}
\begin{IEEEbiography}[{\includegraphics[width=1in,height=1.25in,clip,keepaspectratio]{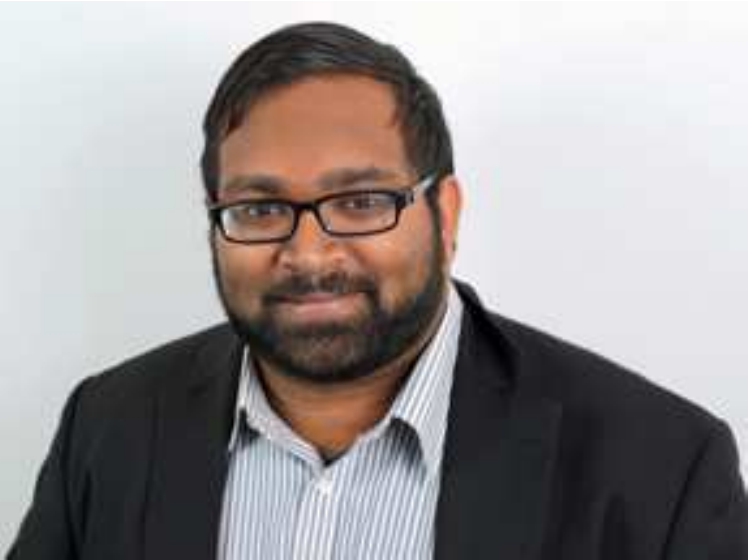}}]{Suren Jayasuriya}
Suren Jayasuriya is an assistant professor at Arizona State University, in the School of Arts, Media and Engineering (AME) and Electrical, Computer and Energy Engineering (ECEE). Before this, he was a postdoctoral fellow at the Robotics Institute at Carnegie Mellon University. Suren received his Ph.D. in ECE at Cornell University in Jan 2017 and graduated from the University of Pittsburgh in 2012 with a B.S. in Mathematics (with departmental honors) and a B.A. in Philosophy. His research interests range from computational cameras, computer vision and graphics, machine learning, sensors, STEAM education, and philosophy. 
\end{IEEEbiography}




\end{document}